\documentclass[10pt,journal,compsoc]{IEEEtran}
\usepackage{epsfig}
\usepackage{graphicx}
\usepackage{amsmath}
\usepackage{amssymb}
\usepackage{amsthm}

\usepackage{comment}
\usepackage{enumitem}
\usepackage{pifont}
\usepackage{url}
\usepackage{booktabs}
\usepackage{makecell}
\usepackage{multirow}
\usepackage{caption}
\usepackage{gensymb}
\usepackage[linesnumbered,lined,boxed,commentsnumbered,ruled]{algorithm2e}
\usepackage[colorlinks,linkcolor=black,anchorcolor=black,citecolor=black,urlcolor=black]{hyperref}
\usepackage{algorithmic}
\usepackage{rotating}
\usepackage[caption=false]{subfig}
\renewcommand{\paragraph}[1]{\noindent\textbf{#1}~~}

\usepackage{bbding}
\usepackage{tabularx}

\ifCLASSOPTIONcompsoc
  \usepackage[nocompress]{cite}
\else
  \usepackage{cite}
\fi

\ifCLASSINFOpdf
\else
\fi

\begin{document}
%
\title{Self-Adversarial Disentangling for Specific Domain Adaptation}
%
%
%

\author{Qianyu~Zhou, 
       Qiqi~Gu, 
       Jiangmiao~Pang, 
       Xuequan~Lu, 
       Lizhuang~Ma
\IEEEcompsocitemizethanks{
\IEEEcompsocthanksitem Q. Zhou, Q. Gu, and L. Ma are with the Department of Computer Science and Engineering, Shanghai Jiao Tong University, Shanghai 200240, China (e-mail: \{zhouqianyu, miemie\}@sjtu.edu.cn, ma-lz@cs.sjtu.edu.cn). \protect
\IEEEcompsocthanksitem J. Pang is with Shanghai AI Laboratory, Shanghai 200232, China (e-mail: pangjiangmiao@gmail.com). 
\IEEEcompsocthanksitem X. Lu is with the School of Information Technology, Deakin University, Victoria 3216, Australia (e-mail: xuequan.lu@deakin.edu.au).

Manuscript received 22 October 2021; revised 19 June 2022 and 21 October 2022; accepted XX December 2022. Date of publication XX 2023; date of current version 18 November 2022.

This work is supported in part by National Key Research and Development Program of China (2019YFC1521104), in part by National Natural Science Foundation of China (72192821, 61972157), in part by Shanghai Municipal Science and Technology Major Project  (2021SHZDZX0102), in part by Shanghai Science and Technology Commission (21511101200), in part by Shanghai Sailing Program (22YF1420300) and in part by Art major project of National Social Science Fund (I8ZD22). 

(Corresponding authors: Lizhuang Ma and Xuequan Lu.)

Recommended for acceptance by XXX.

Digital Object Identifier no. XX.XXXX/XXXXX.2022.XXXXXXX.
}
}

\IEEEtitleabstractindextext{
\begin{abstract}
Domain adaptation aims to bridge the domain shifts between the source and the target domain.
These shifts may span different dimensions such as fog, rainfall, etc.
However, recent methods typically do not consider explicit prior knowledge about the domain shifts on a specific dimension, thus leading to less desired adaptation performance. 
In this paper, we study a practical setting called Specific Domain Adaptation (SDA) that aligns the source and target domains in a demanded-specific dimension.  
Within this setting, we observe the intra-domain gap induced by different domainness (\emph{i.e.,} numerical magnitudes of domain shifts in this dimension) is crucial when adapting to a specific domain.
To address the problem, we propose a novel Self-Adversarial Disentangling (SAD) framework.
In particular, given a specific dimension, we first enrich the source domain by introducing a domainness creator with providing additional supervisory signals.
Guided by the created domainness, we design a self-adversarial regularizer and two loss functions 
to jointly disentangle the latent representations into domainness-specific and domainness-invariant features, thus mitigating the intra-domain gap.  
Our method can be easily taken as a plug-and-play framework and does not introduce any extra costs in the inference time. 
We achieve consistent improvements over state-of-the-art methods in  both object detection and semantic segmentation.

\end{abstract}


}

\maketitle



\IEEEpeerreviewmaketitle
\section{Introduction}\label{sec:introduction}
\IEEEPARstart{O}{ver} the past several years, deep neural networks have brought impressive advances in many computer vision tasks, such as object detection~\cite{girshick2014rich,ren2016faster,chen2017s,focal_loss,shen2019object,cai2019cascade,oksuz2020imbalance,Tan_2022_TPAMI_mirror,zhou222transvod} and semantic segmentation~\cite{fcn, chen2017deeplab,chen2018encoder,zhao2017pspnet,feng2020dmt,lin2017exploring,badrinarayanan2017segnet,night_city}. 
However, the model trained in a source domain will suffer from serious performance degradation when applied to a novel domain, which limits its generalization ability in complicated real-world scenarios. Annotating a large-scale dataset for each new domain is cost-expensive and time-consuming. 
Unsupervised domain adaptation (UDA) emerges, which shows promising results on object detection~\cite{DA-Faster-RCNN,SWDA,MAF,SCDA,GPA,HTCN,ICR-CCR,SCL,hsu2020every,ATF,sindagi2019prior,zhao2020collaborative,ART-PCA,PIT} and semantic segmentation~\cite{ASA,CLANv2,CyCADA, AdaptSegNet,BDL,LTIR,CLAN,SIM, DISE,FDA,STAR,PCEDA,IntraDA,APODA,FADA,DADA,choi2019self,SIBAN,CRST,CBST,Conservative_loss,AdaptPatch,guo2021label},
aiming to reduce the domain shifts between the labeled source domain and the unlabeled target domain.

\begin{figure}[!t] 
\centering
\includegraphics[scale=0.42]{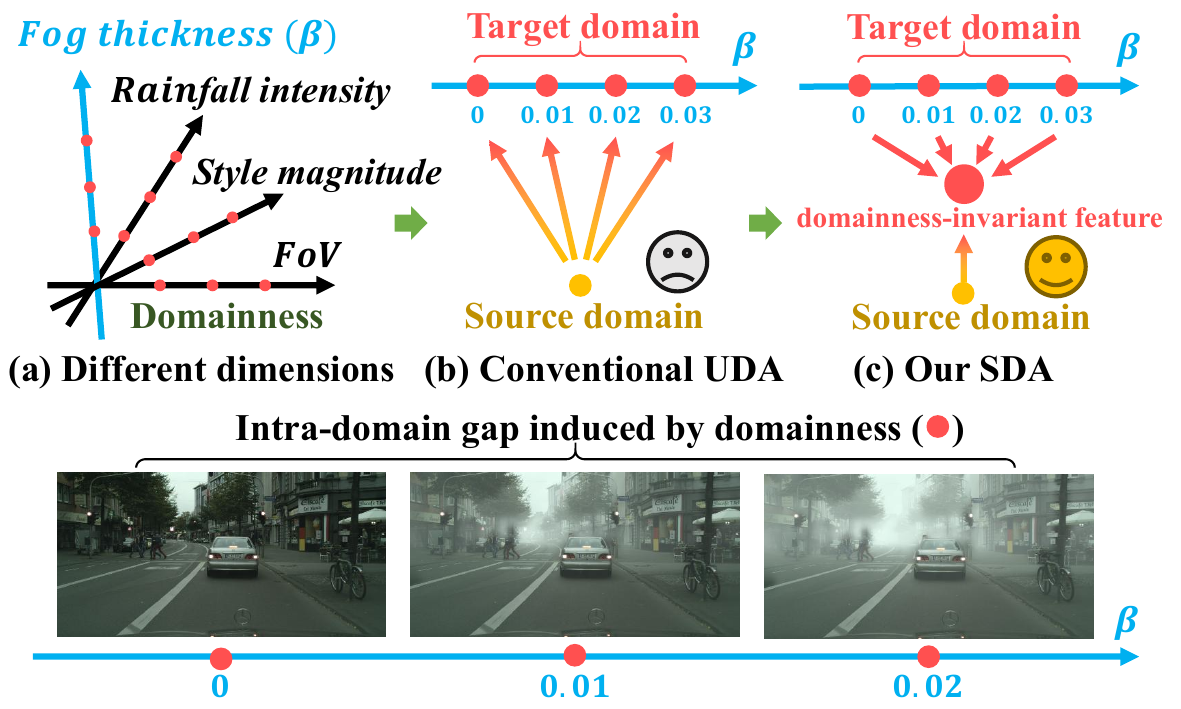}
\caption{
(a) Previous UDA methods do not leverage explicit prior knowledge about the domain shifts on a demand-specific dimension to perform domain adaptation, and (b) they cannot generalize well to a target domain with different unseen domainness ($\beta$). (c) \textit{Our key idea is to learn  domainness-invariant representations for narrowing the intra-domain gap induced by different domainness.} Different \textit{domainness} indicate different numerical magnitudes of domain shifts in a specific domain dimension, \emph{e.g.,} fog thickness. 
}
\label{teaser}
\end{figure}

\begin{table*}{}
\caption{Comparisons of several related settings of UDA, Prior DA, Cross-FOV DA, DG, and Ours (SDA). }
\centering
\label{table-setting}
\resizebox{0.97\textwidth}{!}{%
\begin{tabular}{c|c|c|c|c|c}
\toprule Related  & Labeled  & Unlabeled   & Whether use Target & Domain & Whether Pre-knows\\
settings & Source Domain  & Target Domain & Prior Knowledge & Dimensions & Target Domainness Values \\
\midrule 
Unsupervised Domain Adaptation (UDA)~\cite{GPA,DA-Faster-RCNN,SWDA} & $\checkmark$  & $\checkmark$ &-  &-  & -\\ 
Prior-based Domain Adaptation (Prior-DA)~\cite{sindagi2019prior} & $\checkmark$  & $\checkmark$ & $\checkmark$  & Weather Only & -\\
Cross-FoV Domain Adaptation (Cross-FoV DA)~\cite{PIT} & $\checkmark$  & $\checkmark$ & $\checkmark$ & FoV only & $\checkmark$\\
Domain Generalization (DG)~\cite{li2017domain,ghifary2016scatter} & $\checkmark$  &-   &-   &-  &- \\
\midrule 
Specific Domain Adaptation (SDA) & $\checkmark$  & $\checkmark$  & $\checkmark$ & Arbitrary &- \\
\bottomrule
\end{tabular}
}
\end{table*}
Such domain shifts may span different dimensions, \emph{e.g.,} fog, rainfall, Field of View (FoV), as shown in Fig.~\ref{teaser} (a).
In practical scenarios, there are numerous demands to adapt the model to a specific dimension, \emph{e.g.,} from sunny images to foggy images. For example, in the normal-to-foggy adaptation, there always exist different degrees of fog thickness in the target domain, resulting in low visibility of adverse scenarios and high visibility of clear scenarios. 
However, existing UDA methods can hardly handle such cases effectively. This is mainly because they do not consider any explicit prior knowledge about the domain shifts on a demand-specific dimension. As a result, the model will lack a clear target dimension during the adaptation and will be optimized without especially considering different degrees of fog thickness. Previous UDA models could only show good performance on the seen fog of the target training sets but cannot generalize well to the target testing set with different unseen fog, as shown in Fig.~\ref{teaser} (b). 
This under-constrained training process largely affects the performance when adapting the model to a specific dimension. 

In this work, we use the term, \textit{domainness}, to describe the numerical magnitudes of domain shifts on a specific dimension. For example, in normal-to-foggy/rainy adaptation, different domainness indicates different fog thickness/rainfall intensities of the target domain. 
We observe that such existing domainness with different magnitudes leads to intra-domain gaps, which commonly exist in the same dimension but with different domainness. Such intra-domain gaps are largely overlooked in previous UDA research and could cause the aforementioned performance degradation on the target testing set. As illustrated in Fig.~\ref{teaser} (c), the core idea of our paper is to learn \textit{domainness-invariant features} for well generalizing on specific domains with different domainness. 

In this paper, we refer to the above problem as Specific Domain Adaptation (SDA), a realistic and practical setting for domain adaptation. It targets to align the source and target domains in a demanded-specific dimension, and the model can be broadly applied in real-world applications. For example, in autonomous driving, the models trained on sunny days should have the ability to generalize to specific rainy or foggy scenarios. 
Table \ref{table-setting} lists the differences from related settings. UDA does not consider any prior domain knowledge of the target domain. Prior-DA~\cite{sindagi2019prior} utilized the prior knowledge, \emph{e.g.,} fog or rain, of the target domain, however, it is limited to such weather dimensions only. As for cross-FoV DA~\cite{PIT}, it also utilized the prior knowledge of FoV to enhance the adaptability towards the target domain; however, it requires pre-knowing the existing domainness value of both the source domain and the target domain, which is always unknown in real-world applications. In contrast, 1) our SDA setting can leverage arbitrary (any known or unknown) dimensions of the domain shifts and is no longer limited to weather  or FoV dimension only. 2) Our SDA setting does not require pre-knowing the existing domainness of the target domain, which is more flexible and realistic in practical scenarios. 3) The goal of our SDA setting is to learn  domainness-invariant features, not merely domain-invariant features.

To address the above SDA problem, we present an innovative method, namely Self-Adversarial Disentangling (SAD). From a new perspective, we propose to disentangle the latent representations into domainness-invariant features and domainness-specific features in a specific dimension. 
In comparison to UDA methods that learn domain-invariant features, which do not specifically consider different domainness about the specific domain shifts, \emph{e.g.,} different degrees of fog thickness, we instead learn domainness-invariant feature, which is irrelevant to the domainness magnitude in the target domain. 
The advantage of transferring domainness-invariant features is that we can capture the generalizations across different domainness to narrow down the intra-domain gaps, which is at a more fine-grained level than domain-invariant features. 

Our SAD framework consists of two key components, \emph{i.e.,} Domainness Creator (DC) and Self-Adversarial Regularizer (SAR), for domainness creation and feature disentangling, respectively.  
According to the given domain shift, we firstly enrich the source domain with DC.
It not only diversifies the source domain  but also provides additional supervisory signals for the following feature  disentangling. 
Guided by the domainness, we design the SAR, and introduce a domainness-specific loss and a domainness-invariant loss for SAR to jointly supervise the disentangling of the latent representations into
domainness-specific and domainness-invariant features. 
With the domainness-specific loss, our SAR classifies the predicted domainness with supervisory signals from DC. Penalized by the domainness-invariant loss, our SAR fully learns domainness-invariant representations. Thus, we mitigate the intra-domain gap induced by different domainness. 
 To sum up, our SAD framework works in a disentangling sense, which enables the model to learn domainness-invariant features in an adversarial manner, \emph{i.e.,} two opposite loss functions.

 Our method is applicable and flexible in most real-world cases. We verified the proposed method under various domain dimensions, including cross-fog (Cityscapes~\cite{cordts2016cityscapes} to Foggy Cityscapes~\cite{FoggyCity}, Cityscapes~\cite{cordts2016cityscapes} to RTTS~\cite{RTTS}, Cityscapes~\cite{cordts2016cityscapes} to Foggy Zurich++~\cite{FoggyDriving,FoggyCity}), cross-rain (Cityscapes~\cite{cordts2016cityscapes} to RainCityscapes~\cite{rainy_city}), cross-FoV adaptation (Virtual KITTI~\cite{VKITTI} to CKITTI~\cite{kITTI,cordts2016cityscapes}) and synthetic-to-real adaptation (SIM10K~\cite{sim10k} to Cityscapes~\cite{cordts2016cityscapes}). The target domain has either single or multiple domainness values.  Extensive experiments with analysis prove the impressive generalization abilities of our method.  Without bells and whistles, our method yields remarkable improvements over existing methods in both object detection and semantic segmentation. In particular, we achieve $3.4\% \sim 6.4\%$ gains on synthetic datasets and improvements of up to $2.6\%$ on real datasets.   
We achieve 45.2$\%$ mAP on the widely-used benchmark of Cityscapes~\cite{cordts2016cityscapes} to Foggy Cityscapes~\cite{FoggyCity} and ranked $1_{st}$ in the state-of-the-art UDA benchmark of \href{https://paperswithcode.com/sota/unsupervised-domain-adaptation-on-cityscapes-1}{\textcolor{black}{PaperwithCode}}. 
Our contributions are summarized as follows. 

(1). We study the problem of specific domain adaptation (SDA), a realistic and practical setting for domain adaptation. From a novel perspective, we propose to address the above SDA by disentangling the latent representations into domainness-specific and domainness-invariant representations in a specific dimension.

(2). We present a novel self-adversarial disentangling (SAD) framework by leveraging the explicit prior domain knowledge on a specific dimension to learn the domainness-invariant features. Firstly, we introduce a domainness creator for specifically enriching the source domain and providing explicit supervisory signals. Besides, we design a self-adversarial regularizer to mitigate the intra-domain gaps. We also present one domainness-specific loss and a domainness-invariant loss to facilitate the training.

(3). We conduct comprehensive experiments and analysis on six benchmarks to demonstrate the effectiveness of our proposed method on both object detection and semantic segmentation tasks. It is simple to integrate our method into any existing UDA approaches as a plug-and-play framework that does not introduce any extra costs during the inference phase.

\section{Related Work}
\noindent \textbf{Unsupervised Domain Adaptation.} UDA aims to generalize the model learned from the labeled source domain to another unlabeled target domain. In the field of UDA, a group of approaches have shown promising results in image classification~\cite{zhang2020unsupervised, mancini2019inferring, kouw2019review, 8943120, li2020deep, li2020maximum, rozantsev2018beyond, liang2018aggregating, li2017domain, ghifary2016scatter, courty2016optimal, DANN}. However, most of these methods only work on simple and small classification datasets, and may have quite limited performance in more challenging and higher-structured tasks, \emph{e.g.,} semantic segmentation. Therefore, many researchers study the UDA in object detection~\cite{DA-Faster-RCNN,SWDA,MAF,SCDA,GPA,HTCN,ICR-CCR,SCL,hsu2020every,ATF,sindagi2019prior,zhao2020collaborative,ART-PCA,PIT} and semantic segmentation~\cite{ASA,CLANv2,CyCADA, AdaptSegNet,BDL,LTIR,CLAN,SIM, DISE,FDA,STAR,PCEDA,IntraDA,APODA,FADA,DADA,choi2019self,SIBAN,CRST,CBST,Conservative_loss,AdaptPatch,zhang2019curriculum, luo2021category, sakaridis2020map}.

Despite the gratifying progress, little attention has been paid to perform domain adaptation in a specifically demanded  dimension by introducing any explicit prior knowledge about the domain shifts except ~\cite{sindagi2019prior,PIT}. Prior DA~\cite{sindagi2019prior} is the one of the few works that builds on a similar motivation with us by using the weather-specific prior knowledge obtained from the image formation. However, Prior DA~\cite{sindagi2019prior} only explored the weather prior on the cross-fog and cross-rain scenarios, and it designed a prior-adversarial loss, which acts in a completely different manner from ours. Similarly, Gu \emph{et al.}~\cite{PIT} observed that the Field of View (FoV) gap induces noticeable instance appearance differences between the source and target domains, and presented a Position-Invariant Transform (PIT) to straightforwardly narrow the FoV gap by exploring the prior knowledge about FoV. Nevertheless, these approaches neglected the intra-domain gaps, which commonly exist among the domains in the same dimension but with different domainness. Also, \cite{PIT} needs to pre-know the specific FoV value of both the source and the target domain. In contrast, we study SDA in this paper, which does not require any human annotation about the domainness, and is no longer limited to weather or FoV dimension only.
We deduce that Specific Domain Adaptation (SDA) is a more realistic and practical setting for domain adaptation, which is attracting increasing attention. Following  \cite{sindagi2019prior,PIT}, we know the domain dimension in advance, and conduct the SDA on the cross-fog, cross-rain, cross-FoV scenarios, which is fully fair in experimental comparisons. Table \ref{table-setting} lists the differences from related settings.

\noindent \textbf{Domain Diversification.} Domain Diversification (DD) aims to diversify the source domain to various distinctive domains with random augmentation. Kim \emph{et al.}~\cite{DDMRL} designed a DD-MRL method by using GAN~\cite{GAN} to diversify the source domain. Similarly, DRPC~\cite{DRPC} and LTIR~\cite{LTIR} proposed to diversify the texture of the source images and to learn texture-invariant representations. 
\textit{Our method differs from these methods in several aspects.}  Firstly, they require large computation costs and cannot be trained end-to-end during the adaptation.  Instead, our method is light-weighted and online with a transformation algorithm in DC. Secondly, GAN-based approaches tend to produce artifacts for urban-scene datasets, leading to severe semantic inconsistency. In contrast, we do not use any feature interpolation operation in the reconstruction and merely use a simple yet effective parameter modeling.

\noindent \textbf{Disentangled Learning.} 
Disentangled learning has been widely studied in other communities, \emph{e.g.,} image translation~\cite{huang2018multimodal,lee2018diverse}, few-shot learning~\cite{ridgeway2018learning,scott2018adapted}. 
A few works have recently extended it into domain adaptation by disentangling the latent representations into domain-specific and domain-invariant features to realize effective domain alignment. 
Liu \emph{et al.} proposed a model of cross-domain representation disentanglement (CDRD)~\cite{liu2018detach} based on the GAN~\cite{GAN} framework. Chang \emph{et al.} designed a domain invariant structure extraction (DISE) framework~\cite{DISE} to disentangle the latent encodings into the domain-invariant structure and domain-specific texture representations for domain-adaptive semantic segmentation. Wu \emph{et al.}~\cite{IID} presented a progressive disentanglement to learn the instance-invariant features in domain adaptive object detection. 
Nevertheless, these methods all perform the domain disentanglement to learn domain-invariant features instead of domainness-invariant features, and it
can hardly capture the generalizations across different domainness within the same target domain to narrow down the intra-domain gap.

\begin{figure*}[t]
\centering
\includegraphics[scale=0.42]{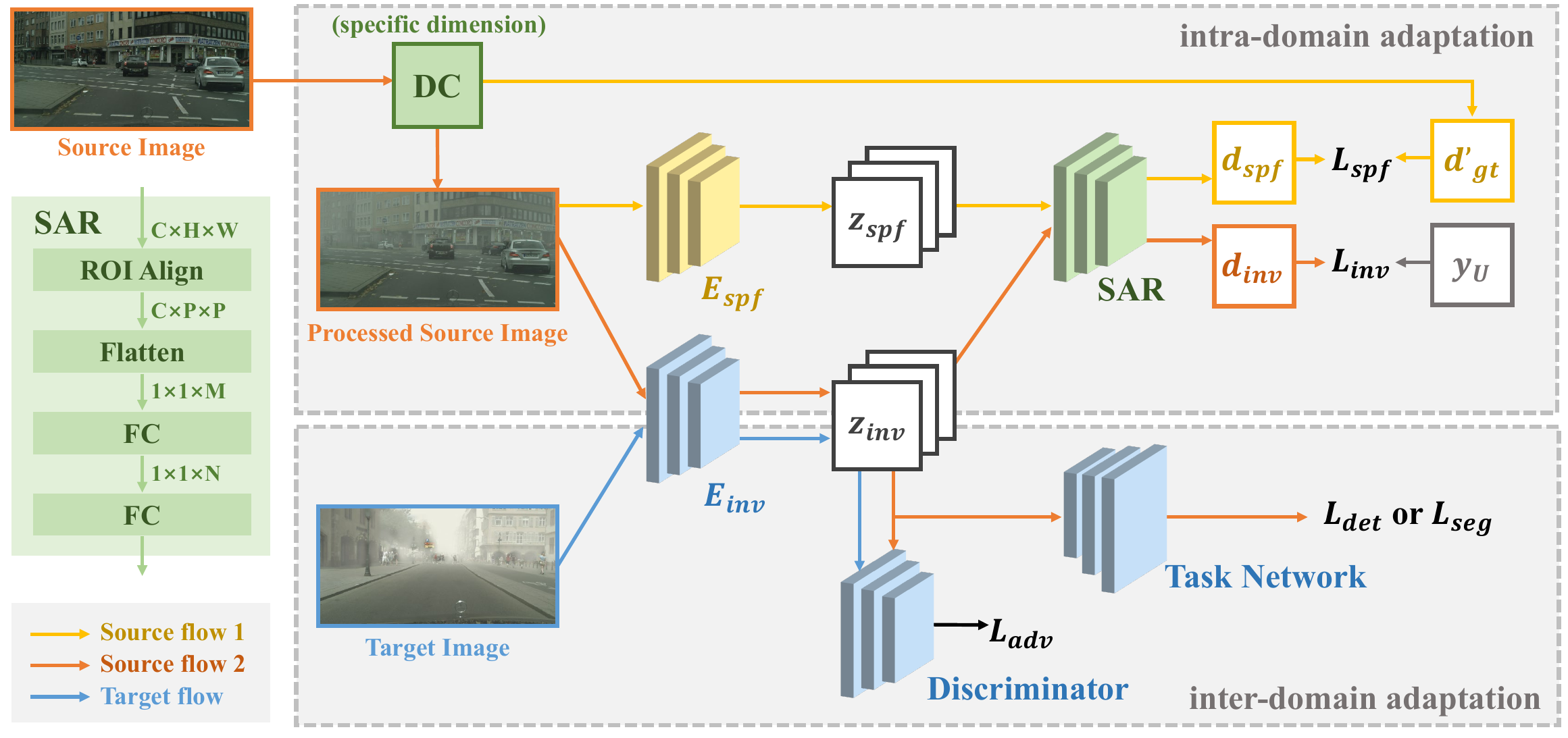}
\caption{Overview of the proposed Self-Adversarial Disentangling (SAD) framework for specific domain adaptation (SAD). Our Domainness Creator (DC) not only generates a diversified source image with random domainness, but also provides additional supervisory signals $d'_{gt}$ for guiding the feature disentangling. The encoder $E_{spf}$ and $E_{inv}$ are to extract the domainness-specific representations $z_{spf}$ and the domainness-invariant representations $z_{inv}$, respectively. With the guidance of the generated domainness, $E_{spf}$, $E_{inv}$ and SAR (Self-Adversarial Regularizer) work in an adversarial manner, \emph{i.e.,} two opposite loss functions, to disentangle the latent representations into $z_{spf}$ and  $z_{inv}$ (Best viewed in color). 
}
\label{fig:framework}
\end{figure*}

\section{Methodology}
We focus on the problem of Specific Domain Adaptation (SDA) in both object detection and semantic segmentation, where we have access to the source data $X_S$ with labels $Y_S$ and the target data $X_T$ without labels.  Fig.~\ref{fig:framework} shows the overview of our framework. Our core idea is to disentangle  the  latent  representation into  domainness-invariant feature and domainness-specific feature in a specific  dimension, thus bridging intra-domain and inter-domain gap. The target domain has either single or multiple domainness.

\subsection{Domainness Creator}
\label{sec:domainness creator}
\textbf{Definitions of domainness.} \textit{Domainness describes the numerical magnitudes of domain shifts on a specific dimension, which induces the intra-domain gap of the target domain.} 
Taking the fog/rain or FoV dimension as an example,  if we normalize the range of domainness to 0\% $\sim$ 100\%, 100\% domainness means the largest fog thickness/rainfall intensities/FoV values existing in the target domain. \textit{Also, domainness reflects the strength of the augmentation during the domainness creation}. 0\% domainness denotes that the source domain remains its original fog thickness/rainfall intensities/FoV values, and 100\% domainness indicates that the source domain's fog thickness/rainfall intensities/FoV values will be fully replaced by the target domain's largest fog thickness/rainfall intensities/FoVs. Fig. \ref{fig:dc_results} illustrates three different domainness values of style magnitudes/fog thickness/rainfall intensities/FoVs.
As such, if we control the value of domainness during the domainness creation phase, we would get continuous intermediate domains between the source and the target domain. 

\noindent \textbf{Overview of Domainness Creator.} To diversify the source images given the domain dimension,
we design Domainness Creator (DC) as a transformation algorithm. 
DC receives a source image $X_S$ as the input and outputs a processed image $\tilde{X}_{S}$ by adding a random domainness in a specific dimension.
 Meanwhile, DC provides a  supervisory signal, \emph{i.e.,}, the category label $d'_{gt}$ of the domainness value $d_{gt}$,  for guiding the self-adversarial learning.  Due to the  variations of domainness values enabled by DC, a model trained on the  domainness-diversified dataset will be able to learn the domainness-invariant representations for feature alignment. $d_{gt}$ is a number, \emph{e.g.,} $FoV_x$ is $40^{\circ}$. $FoV_x$ denotes the FoV in the $x$ axis.

\begin{figure}[!t]
\centering
\includegraphics[scale=0.28]{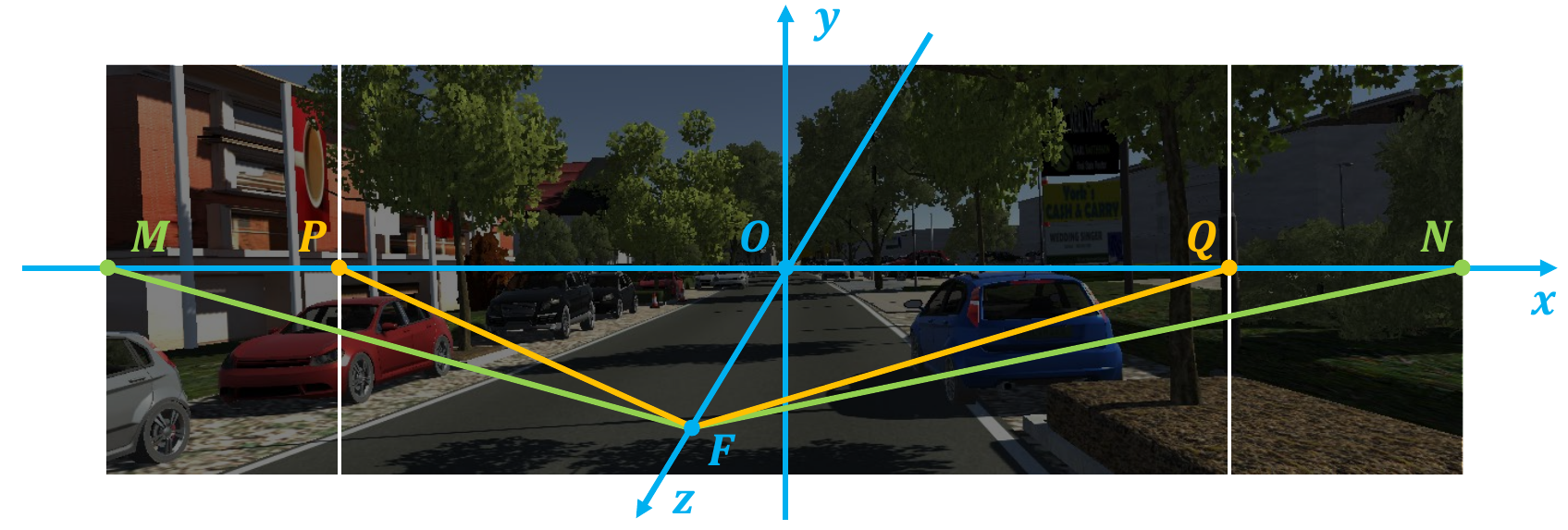}
\caption{The $FoV_{x}$ transform. $O$ is the optical center of the camera and $F$ is the focal point. $OF$ is the focal length, $MN$ and $PQ $ represent the original width and the new width before and after the transformation, respectively.  $FoV_{x}$ is reduced from $\angle MFN$ to $\angle PFQ$ after the process. }
\label{fig:crop}
\end{figure}

\noindent \textbf{Example of Domainness Creator in FoV dimension.} Taking FoV dimension as an example, we show the process of FoV transformation given a selected $FoV_{x}$ in Fig.~\ref{fig:crop}, where $O$ is the optical center of the camera and $F$ is the focal point. $OF$ denotes the focal length. $MN$ and $PQ$ represent the original width and the new width before and after the transformation: 
\begin{align}
\tilde{X}_{S} = DC(X_S), FOV_x = \angle MFN \rightarrow \angle PFQ
\end{align}
where $FoV_{x}$ is  reduced from $\angle MFN$ to $\angle PFQ$ during the process and the generated domainness is denoted as $d_{gt}=\angle PFQ$. 
If the dimension is fog or rain, we follow the algorithms in \cite{FoggyCity,rainy_city} for diversification. If we do not have prior knowledge about the domain shifts, our DC will employ the style of the target domain by default, and adopt \cite{FDA} for diversification, which also provides a universal solution when tackling unknown domain shifts.

\vspace{-2mm}

\subsection{Self-Adversarial Regularizer}
\label{sec: sar}

Guided by the generated domainness, SAR is designed to disentangle the latent representations into the domainness-specific feature $z_{spf}$ and the domainness-invariant feature $z_{inv}$, in order to mitigate the intra-domain gap. 
$E_{spf}$ and $E_{inv}$ denote the domainness-specific encoder and domainness-invariant encoder. 
The dimensions of $z_{spf}$ and $z_{inv}$ are both $C*H*W$, where $C$ is 19/11 for segmentation, and 512/1024 for detection, respectively.

\paragraph{Intra-domain adaptation.}
As shown in Fig. \ref{fig:framework}, the processed image $\tilde{X}_{S}$ is fed into the encoder $E_{spf}$ and $E_{inv}$ to get the latent feature map $z_{spf}$ and $z_{inv}$. Either $z_{spf}$ or  $z_{inv}$ is forwarded into SAR to get the domainness value $d_{spf}$ and $d_{inv}$ for once.  
SAR is supervised by the designed domainness-specific loss $\mathcal{L}_{spf}$ and  domainness-invariant loss $\mathcal{L}_{inv}$ together (see below for the design of these two losses). With the former loss $\mathcal{L}_{spf}$, our SAR could classify the predicted domainness $d_{spf}$ with supervisory signals $d_{gt}$ from DC.  Penalized by the latter loss $\mathcal{L}_{inv}$, our SAR fully learns domainness-invariant representations, thus mitigating the intra-domain gap induced by different domainness. 
In essence, the two branches (yellow and orange branch in Fig. \ref{fig:framework}) are complementary and the SAR module work in an \textit{self-adversarial} manner (\emph{i.e.,} two opposite losses) to perform the specific domain adaptation. We illustrate the network details and the two loss fuctions below.

\paragraph{Network architecture of SAR.}
Note that we use the same SAR architecture for both object detection and semantic segmentation. SAR only takes one feature map $z_{spf}$ or $z_{inv}$ at a time as input. After that, we downsample the whole feature map to predict domainness value, and then flatten the downsampled feature map. Then, after two fully-connected layers with a relu activation, we get the domainness $d_{spf}$ or $d_{inv}$, as shown in Fig.~\ref{fig:framework}. In practice, we use ROI Align~\cite{mask-rcnn} to downsample the \emph{whole} feature map to predict domainness value.  
We discretize continuous domainness values into $N$ numbers (representing $N$ ranges)  for better experimental results.
$d_{spf}$, $d_{inv}$ are one-hot vectors with $N$ dimensions. $y_{U}$ is a $N$ dimensional vector of the uniform distribution.

\paragraph{Domainness-specific loss.}
 On the one hand, with the generated domainness label $d'_{gt}$ as a supervisory signal, SAR needs to enhance its discriminativity for classifying the diversified images with different domainness more accurately, which is illustrated in the yellow branch of Fig.~\ref{fig:framework}.
The predicted domainness is $d_{spf}=SAR(E_{spf}(\tilde{X}_{S}))$.
We define the domainness-specific loss $\mathcal{L}_{spf}$ as a cross-entropy loss for optimizing the features from the encoder $E_{spf}$:
\begin{align}
\label{eq:domainness-specific}
  \mathcal{L}_{spf} = -\sum\limits_{i=1}^{N} d'^{i}_{gt} log(d^{i}_{spf}),
\end{align}
where $d'_{gt}$ is used as the one-hot label vector of domainness value $d_{gt}$ and $d_{spf}$ is the predicted domainness of SAR. 

\paragraph{Domainness-invariant loss.}
 On the other hand, SAR needs to maximize the discrepancy between the domainness-invariant feature $z_{inv}$ and the domainness-specific feature $z_{spf}$. In this branch (the orange branch of Fig.~\ref{fig:framework}), the predicted domainness is: $d_{inv}=SAR(E_{inv}(\tilde{X}_{S}))$. We define the domainness-invariant loss $\mathcal{L}_{inv}$ as the KL-divergence between the predicted domainness $d_{inv}$ and a uniform distribution $y_{U}$: 
\begin{equation}
    \begin{aligned}
    \label{eq:domainness-inv}
&z_{inv} \sim E_{inv}\left(\tilde{X}_{S}\right) =q_{S}\left(d_{inv} \mid \tilde{X}_{S}\right), \\
&\mathcal{L}_{inv} = KL\left(q_{S}\left(d_{inv} \mid \tilde{X}_{S}\right) \| p\left(x\right) \right),
\end{aligned}
\end{equation}
where $x$ is sampled from a uniform  distribution $y_{U}$, $p(x)=\frac{1}{\Psi}$ is the probability of $x$, and $\Psi$ is the number of domainness.
$q_{S}$ denotes the distribution of domainness $d_{inv}$. As such, the predicted domainness $d_{inv}$ is desired to be equal to the average, indicating that SAR is learning to degenerate its capabilities in discriminating the domainness value. 

By jointly minimizing the domainness-invariant loss $\mathcal{L}_{inv}$ and the domainness-specific loss $\mathcal{L}_{spf}$ in two inverse directions, our SAD framework works in an disentangling sense by disentangling the latent feature into domainness-specific feature $z_{inv}$ and domainness-invariant feature $z_{inv}$. Therefore,
SAR can fully learn the domainness-invariant features in an adversarial manner, which capture the generalizations across different domainness, thus narrowing the intra-domain gap.

\noindent \textbf{Remark 1:} \textit{Whether the parameters of $E_{spf}$ and $E_{inv}$ are shared or not.} $E_{spf}$ and $E_{inv}$ are two encoders that use the same architecture but do not share the weights as they are penalized by different loss functions. The former is penalized by $L_{spf}$, and the latter is under the guidance of $L_{inv}$, $L_{adv}$ (adversarial loss, Eq. \eqref{eq:ladv}) and $L_{task}$ (task loss, Eq. \eqref{eq:task}).

\noindent \textbf{Remark 2:} \textit{Comparing with GAN architecture.} 
Existing GAN-based architectures (multi domain-invariant representation learning) utilized the multi-domain discriminators~\cite{DDMRL,CIDA} to distinguish the domainness (they called domain index in their work). In the adversarial framework, these discriminators are not actually predicting the domainness $d_{inv}$, but making the latent encodings $z_{inv}$ unable to predict $d_{inv}$. Due to the fact that it is trained in an adversarial way, the encoder will transform the input $X$ before outputting encoding $z_{inv}$, thereby removing the information related to domainness $d_{inv}$. 
However, the encoder can not fully learn the domainness-invariant feature due to the lack of prior knowledge about the domain shifts. 
\textit{In comparison, our proposed framework acts in a completely different manner.} Firstly, we use two separate encoders $E_{spf}$ and $E_{inv}$ instead of one encoder, the former for extracting the domainness-specific feature $z_{spf}$ and the latter for extracting the domainness-invariant features $z_{inv}$. Secondly, with the guidance of the generated domainness $d'_{gt}$ as supervisory signals, our SAR is truly reconstructing the domainness, aiming to distinguish the domainness accurately; Thirdly,  our SAD framework works in two opposite directions in a disentangling sense, which enables the model to learn domainness-invariant features to alleviate the intra-domain gap.

\subsection{Overall Objective and Optimization}
\label{END-TO-END}
 In this section, we will briefly introduce the inter-domain adaptation, the task loss and formulate an overall loss function for the training process. Then we will explain the inference phase.

\begin{algorithm}[htb!]
	\caption{Self-Adversarial Disentangling}
	\label{algorithm 1}
	\KwIn{source domain $\{X_S, Y_S\}$ and target domain $\{X_T\}$, disentangled feature extractors $E_{spf}$, and $E_{inv}$, self-adversarial regularizer $SAR$, task model $T$, and inter-domain adaptor $D$.
	}
	\KwOut{domainnes-invariant feature extractor $E_{inv}$ and task model $T$.}
	Initialize network parameters $\theta$ randomly\;
	\For{$t \leftarrow 1$ \KwTo $N$}{
		Sample a mini-batch from $\{X_S, Y_S\}$ and $\{X_T\}$\;
		\textbf{Domainness Creation:}\\
		Pre-process the source image $X_S$ in a specified dimension and get the diversified image $\tilde{X}_S$\;
		Generate a random domainness value and assign a category as the domainness label $d'_{gt}$\; 
        \textbf{Feature Disentangling:}\\
        Estimate the predicted domainness:
        $d_{spf}, d_{inv} = SAR(E_{spf}(\tilde{X}_S)), SAR(E_{inv}(\tilde{X}_S))$\;
		Compute domainness-specific loss $\mathcal{L}_{spf}$ (Eq. (\ref{eq:domainness-specific}))\;
        Reckon domainness-invariant loss $\mathcal{L}_{inv}$ (Eq. (\ref{eq:domainness-inv}))\;
        \textbf{Inter-domain Adaptation:}\\
        Calculate a common inter-domain adaptation loss, \emph{e.g.,} $\mathcal{L}_{adv}$ (Eq. (\ref{eq:ladv})), and the task loss $\mathcal{L}_{task}$ with the source supervision (Eq. (\ref{eq:task}))\;
		Compute $\bigtriangledown_{\theta}\mathcal{L}_{total}$ by back-propagation (Eq. (\ref{eq:total}))\;
		Perform stochastic gradient descent on $\theta$\;
	}
	return $E_{inv}$, $T$
\end{algorithm}


\paragraph{Inter-domain adaptation.}
Without loss of generality, we employ an adversarial framework~\cite{DANN} for the inter-domain adaptation. 
As shown in Fig. \ref{fig:framework}, the processed source images $\tilde{X}_{S}$ and ${X}_{T}$ are fed into the encoder $E_{inv}$. Then, $E_{inv}$ is encouraged to learn $z_{inv}$. The latent encodings should confuse a domain discriminator $D$ in distinguishing the features extracted between the source and target domains. This is achieved by min-maximizing an adversarial loss: 
\begin{equation}
    \begin{aligned}
\label{eq:ladv}
\mathcal{L}_{a d v}=&-\mathbb{E}_{\mathbf{x} \sim p\left(\tilde{\mathbf{x}}_S\right)}[\log (D(E_{inv}(\mathbf{x})))] \\
&-\mathbb{E}_{\mathbf{x} \sim p\left(\mathbf{x}_{T}\right)}[\log (1-D(E_{inv}(\mathbf{x})))],
\end{aligned}
\end{equation}


\noindent \textbf{Task loss.} In this work, taking Faster-RCNN~\cite{ren2015faster} as an example of the training model, we use Region Proposal Network (RPN) to generate Region of Interests (RoIs). It then localizes and classifies the regions to obtain semantic labels and locations. The task network is optimized with a multi-task loss function:
\begin{align}
\label{eq:task}
  \mathcal{L}_{task} = \mathcal{L}_{rpn} + \lambda_{cls} \mathcal{L}_{cls} + \lambda_{reg} \mathcal{L}_{reg} ,
\end{align}
where the RPN loss $\mathcal{L}_{rpn}$, classification loss $\mathcal{L}_{cls}$ and regression loss $\mathcal{L}_{reg}$ remain the same as \cite{ren2015faster}. The
loss weights $\lambda_{cls}$ and $\lambda_{reg}$ are set to 1.0 by default.

\noindent \textbf{Total loss.} During training, all the models are jointly trained with the backbone in an end-to-end manner. The total loss $L_{total}$ is the weighted sum
of the aforementioned losses:
\begin{align}
\label{eq:total}
  \mathcal{L}_{total} = \mathcal{L}_{task} + \lambda_{adv} \mathcal{L}_{adv} +\mathcal{L}_{inv} + \lambda_{spf} \mathcal{L}_{spf},
\end{align}
where $\lambda_{adv}$ and $\lambda_{spf}$ are the weighting coefficients for the loss $\mathcal{L}_{adv}$ and $\mathcal{L}_{spf}$, respectively. We use the original weighting ratio in \cite{DA-Faster-RCNN,SWDA,SCL,GPA,AdaptSegNet,CLAN,SIM} to balance $\mathcal{L}_{task}$ and $\mathcal{L}_{adv}$.  The overall training algorithm is described in Algorithm \ref{algorithm 1}.

\noindent \textbf{Inference phase.} In the inference phase, we only need a domainness-invariant encoder $E_{inv}$ with a task network $T$ to make predictions. In other words, all other modules including DC, SAR and $E_{spf}$ are removed in the inference stage, leading to no extra costs in prediction.  Besides, our method can be plugged into various existing cross-domain detection/segmentation methods.  Thus, our framework is flexible and generalizable, and it does not depend on specific UDA frameworks for feature alignment.

\begin{table*}
\caption{
Comparison results for object detection in (a,b) cross-fog, (c) cross-FoV and (d) cross-rain scenarios. 
}
\centering
\resizebox{\textwidth}{!}{%
\subfloat[Cross-fog adaptation on Cityscapes to Foggy Cityscapes (single-domainness).\label{table:ob_comb:benchmark}]{
\begin{tabular}{c|c|cccccccc|c}
\toprule
Methods & backbone & person & rider & car & truck & bus & train & motor & bicycle & mAP \\
\toprule
Source Only~\cite{ren2016faster}& VGG16 & 26.9 & 38.2 & 35.6 & 18.3 & 32.4 & 9.6 & 25.8 & 28.6 & 26.9 \\
BDC-Faster \cite{SWDA} & VGG16  & 26.4 & 37.2 & 42.4 & 21.2 & 29.2 & 12.3 & 22.6 & 28.9 & 27.5 \\
SCDA~\cite{SCDA} & VGG16 &33.5 &38.0 &48.5 &26.5 &39.0 &23.3 &28.0 &33.6 &33.8 \\
DD-MRL\cite{DDMRL} & VGG16 & 30.8 & 40.5 & 44.3 & 27.2 & 38.4 & 34.5 & 28.4 & 32.2 & 34.6  \\
SWDA\cite{SWDA} & VGG16 & 29.9 & 42.3 & 43.5 & 24.5 & 36.2 & 32.6 & 30.0 & 35.3 & 34.3  \\
ICR-CCR~\cite{ICR-CCR} &VGG16 & 32.9 & 43.8 & 49.2 & 27.2 & 45.1 & 36.4 & 30.3 & 34.6 & 37.4 \\
ATF~\cite{ATF}&VGG16 &34.6 &47.0 &50.0 &23.7 &43.3 &38.7 &33.4 &38.8 &38.7 \\
 Prior DA~\cite{sindagi2019prior} &VGG16 &36.4 &47.3 &51.7 &22.8 &47.6 &34.1 &36.0 &38.7 &39.3\\
 RPN-PA~\cite{zhang2021rpn} &VGG16 & 33.3 & 45.6 & 50.5 & 30.4 & 43.6 & 42.0 & 29.7 & 36.8 & 39.0 \\
MeGA-CDA~\cite{vs2021mega} &VGG16 & 37.7 & 49.0 & 52.4 & 25.4 & 49.2 & 46.9 & 34.5 & 39.0 & 41.8 \\
UMT~\cite{deng2021unbiased} &VGG16 & 33.0 & 46.7 & 48.6 & 34.1 & 56.5 & 46.8 &  30.4 & 37.3 & 41.7 \\
VDD~\cite{wu2021vector} &VGG16 & 33.4 & 44.0 & 51.7 & 33.9 & 52.0 & 34.7 & 34.2 & 36.8 & 40.0 \\
\midrule
DA-Faster~\cite{DA-Faster-RCNN}& VGG16 & 25.0 & 31.0 & 40.5 & 22.1 & 35.3 & 20.2 & 20.0 & 27.1 & 27.6 \\
\textbf{Ours (with DA-Faster~\cite{DA-Faster-RCNN})}& VGG16 &31.8 &43.8 &51.7 &18.0 &30.1 &10.3 &29.0 &34.9 &\textbf{31.2}\\
GPA~\cite{GPA}&VGG16 &36.5 &45.2 &55.6 &25.4 &45.1 &18.0 &35.0 &39.3 &37.5 \\
\textbf{Ours (with GPA~\cite{GPA})}& VGG16 &37.2 &48.1 &59.1 &28.4 &50.7 &44.0 &36.7 &38.6 &\textbf{42.9}\\
Oracle (Target Only) & VGG16 &36.2 &46.5 &52.8 &34.0 &53.1 &40.2 &36.0 &36.4 &41.9\\
\midrule
Source Only~\cite{ren2016faster}  &ResNet50 & 36.9 &36.1 & 44.5 &21.7 &32.3 &9.2 &21.5 &32.4 &28.3 \\
DA-Faster\cite{DA-Faster-RCNN} & ResNet50 & 29.2 & 40.4 & 43.4 & 19.7 & 38.3 & 28.5 & 23.7 & 32.7 & 32.0  \\
MAF~\cite{MAF}&ResNet50 &28.2 &39.5 &43.9 &23.8 &39.9 &33.3 &29.2 &33.9 &34.0 \\
DD-MRL~\cite{DDMRL}&ResNet50 & 31.8 & 40.5 & 51.0 & 20.9 & 41.8 & 34.3 & 26.6 & 32.4 & 34.9 \\
SWDA\cite{SWDA} & ResNet50 & 31.8 & 44.3 & 48.9 & 21.0 & 43.8 & 28.0 & 28.9 & 35.8 & 35.3   \\
SCDA\cite{SCDA}  & ResNet50 & 33.8 & 42.1 & 52.1 & 26.8 & 42.5 & 26.5 & 29.2 & 34.5 & 35.9  \\
MTOR~\cite{MTOR} &ResNet50 & 30.6 & 41.4 & 44.0 & 21.9 & 38.6 & 40.6 & 28.3 & 35.6 & 35.1 \\
IID~\cite{IID} & ResNet50 & 32.8 & 44.4 & 49.6 & 33.0 & 46.1 & 38.0 & 29.9 &35.3 &38.6 \\
\midrule
GPA~\cite{GPA}&ResNet50 &32.9 &46.7 &54.1 &24.7 &45.7 &41.1 &32.4 &38.7 &39.5 \\
\textbf{Ours (with GPA~\cite{GPA})}& ResNet50 &\textbf{38.3} & \textbf{47.2} & \textbf{58.8} & \textbf{34.9} & \textbf{57.7} & \textbf{48.3} & \textbf{35.7} & \textbf{42.0} & \textbf{45.2} \\
Oracle (Target Only) & ResNet50 &37.7 &44.1 &58.0 &37.7 &55.9 &47.1 &33.6 &36.5 &43.8\\
\bottomrule
\end{tabular}
}
}
\resizebox{0.63\textwidth}{!}{%
\subfloat[Cross-fog adaptation on Cityscapes to RTTS (multi-domainness).\label{table:ob_comb:benchmark2_rtts}]{
\begin{tabular}{c|c|ccccc|c}
\toprule
Methods& backbone  & car & bus & person & motor & bicycle & mAP \\
\toprule
Source Only~\cite{ren2016faster} & VGG16 & 39.8 & 11.7 &  46.6  & 19.0 & 37.0 & 30.9  \\
DCPDN~\cite{zhang2018densely}  & VGG16 & 39.5 & 12.9& 48.7 & 19.7 & 37.5 & 31.6 \\
Grid-Dehaze~\cite{liu2019griddehazenet}  & VGG16 & 25.4 & 10.9 & 29.7 & 13.0 & 21.4 & 20.0 \\
    DA-Faster~\cite{DA-Faster-RCNN}  & VGG16 & 43.7 & 16.0 & 42.5 & 18.3 & 32.8 & 30.7 \\
SWDA~\cite{SWDA}  & VGG16 & 44.2 & 16.6  & 40.1 & 23.2 & 41.3 & 33.1 \\
\midrule
\textbf{Ours (with~\cite{DA-Faster-RCNN})} & VGG16  & 45.0 & 15.9 & 42.0  & 22.2 & 38.4 & 32.7 \\
\textbf{Ours (with~\cite{SWDA})}  & VGG16 & \textbf{47.0} & \textbf{16.6} &41.5 &\textbf{27.2} &\textbf{43.2} & \textbf{35.1} \\
\midrule
Oracle (Target Only) & VGG16 &63.5  &48.9  &61.0 &58.4 &56.6 &57.7  \\
\bottomrule
\end{tabular}
}
}
\quad
\resizebox{0.35\textwidth}{!}{%
\subfloat[Cross-FoV adaptation on Virtual KITTI to CKITTI (multi-domainness) \label{table:ob_comb:v2c_det}]{
\begin{tabular}{c|c|c|c} 
\toprule
Comparisons & backbone & Car AP & Gain\\
\toprule
Source Only~\cite{ren2016faster} & VGG16 & 42.9 & - \\
\midrule
DA-Faster~\cite{DA-Faster-RCNN} & VGG16 & 45.1 & \multirow{2}{*}{\textbf{2.6}}\\
\textbf{ Ours (with~\cite{DA-Faster-RCNN})} & VGG16 & \textbf{47.7} & \\
\midrule
 SWDA~\cite{SWDA} & VGG16 & 49.0 & \multirow{2}{*}{\textbf{1.7}}\\
\textbf{ Ours (with~\cite{SWDA})} & VGG16  & \textbf{50.7} &  \\
\midrule
SCL~\cite{SCL} & VGG16 & 49.5 & \multirow{2}{*}{\textbf{1.8}}\\
\textbf{ Ours (with~\cite{SCL})} & VGG16  & \textbf{51.3} & \\
 \midrule
 Oracle (Target Only) & VGG16  & 64.0 & - \\
\bottomrule
\end{tabular}
}
}
\resizebox{\textwidth}{!}{%
\subfloat[Cross-rain adaptation on Cityscapes to RainCityscapes (multi-domainness).\label{table:ob_comb:benchmark}]{
\begin{tabular}{c|c|ccccccc|c|c}
\toprule Methods & backbone & person & rider & car & truck & bus & motor & bicycle & mAP & Gain\\
 \toprule
 DA-Faster~\cite{DA-Faster-RCNN} & VGG16 & 22.9 & 55.2 & 43.4 & 3.9 & 58.8 & 15.2 & 30.0 & 32.8 & \multirow{2}{*}{\textbf{6.4}}\\
\textbf{Ours (with DA-Faster~\cite{DA-Faster-RCNN})} & VGG16 & \textbf{26.3} & \textbf{60.1} & \textbf{52.6} & \textbf{13.0} & \textbf{60.3} & \textbf{27.0} & \textbf{34.9} & \textbf{39.2} \\
\midrule
SWDA~\cite{SWDA} & VGG16 & 23.8 & 52.1 & 46.4 & 9.6 & 68.2 & 16.0 & 32.8 & 35.6 & \multirow{2}{*}{\textbf{3.4}}\\
\textbf{Ours (with SWDA~\cite{SWDA})} & VGG16 & \textbf{25.9} & \textbf{56.0} & \textbf{52.5} & 8.1 & \textbf{56.0} & \textbf{29.4} & \textbf{33.1} & \textbf{39.0} &\\
\midrule
SCL~\cite{SCL} & VGG16 & 27.0 & 57.9 & 50.3 & 10.0 & 67.9 & 13.9 & 33.9 & 37.3 & \multirow{2}{*}{\textbf{4.2}}\\
\textbf{Ours (with SCL~\cite{SCL})} & VGG16 & \textbf{29.3} & \textbf{61.0} & \textbf{52.7} & \textbf{19.2} & \textbf{68.2} & \textbf{26.2} & \textbf{34.1} & \textbf{41.5} &\\
\midrule
Oracle (Target Only) & VGG16 &26.0  &55.3  &52.1  &3.7  &47.9  &4.5  &28.0  &31.1  &-\\
\bottomrule
\end{tabular}
}
}
\label{table:ob_comb}
\end{table*}

\section{Experiments}
In this section, we describe experimental setups in Section~\ref{sec:datasets} and implementation details in Section~\ref{sec:implementation details}. Then, we demonstrate the effectiveness of our framework on domain adaptive object detection (Section~\ref{sec:DAOD}) under various domain dimensions, including cross-fog adaptation, cross-rain adaptation, cross-FoV adaptation, and synthetic-to-real adaptation.  In addition, we extend our method to the domain adaptive semantic segmentation (Section~\ref{sec:DASS}) to verify its scalability and applicability. Finally, we conduct ablation studies and visualizations to 
show the contribution of each component of our method in Section~\ref{sec:ablation}.
 
\subsection{Datasets}
\label{sec:datasets}
\paragraph{Cityscapes $\rightarrow$ Foggy Cityscapes.} This is a widely-used benchmark for cross-domain object detection. Cityscapes ~\cite{cordts2016cityscapes} is a dataset focused on autonomous driving, which consists of 2,975 images in the training set, and 500 images in the validation set. 
Foggy  Cityscapes~\cite{FoggyCity} is a synthetic foggy dataset which simulates fog on real scenes. 
The annotations and data split in Foggy Cityscapes are inherited from Cityscapes.

\begin{table*}{}
\caption{Cross-FoV adaptation of semantic segmentation from Virtual KITTI to CKITTI (multi-domainness). }
\label{table:v2c_seg}
\centering
\resizebox{\textwidth}{!}{%
\begin{tabular}{c|c|ccccccccccc|c|c}
\toprule
Method    &backbone            & \begin{turn}{90}road\end{turn} & \begin{turn}{90}building\end{turn} & \begin{turn}{90}pole\end{turn} & \begin{turn}{90}light\end{turn} & \begin{turn}{90}sign\end{turn} & \begin{turn}{90}vegetation\end{turn} & \begin{turn}{90}terrain\end{turn} & \begin{turn}{90}sky\end{turn}  & \begin{turn}{90}car\end{turn} & \begin{turn}{90}truck\end{turn} & \begin{turn}{90}guard rail\end{turn}  & \begin{turn}{90}\textbf{mIoU}\end{turn}
& \begin{turn}{90}\textbf{Gain}\end{turn}\\ 
 \midrule
AdaptSegNet~\cite{AdaptSegNet} &ResNet101 &88.0	&80.6	&11.1	&17.4	&28.4	&80.3	&29.2	&85.2	&82.1	&29.7	&27.5	&50.8 & \multirow{2}{*}{\textbf{1.8}}\\
\textbf{Ours (with AdaptSegNet~\cite{AdaptSegNet})} &ResNet101 &\textbf{88.4}	&\textbf{81.0} &9.7	&\textbf{18.9}	&\textbf{30.5}	&\textbf{80.9}	&\textbf{39.1}	&\textbf{86.2}	&\textbf{83.6}	&\textbf{32.6}	&\textbf{27.5}	&\textbf{52.6} &\\
\midrule
CLAN~\cite{CLAN} & ResNet101 &88.2  &80.0  &6.0	 &17.9	 &26.7	 &79.3	 &36.1	 &85.7	 &82.4	 &28.5	 &12.3	 &49.4 &\multirow{2}{*}{\textbf{1.1}}\\
\textbf{Ours (with CLAN~\cite{CLAN})} & ResNet101 &88.1	&79.9	&\textbf{9.9}	&\textbf{19.6}	&25.3	&\textbf{80.2}	&\textbf{38.5}	&\textbf{85.9}	&\textbf{82.5}	&\textbf{29.2}	&\textbf{16.4}	&\textbf{50.5} &\\
\midrule
SIM~\cite{SIM} & ResNet101 &87.3	&81.2	&16.3	&16.1	&28.3	&81.6	&37.6	&87.2	&82.6	&29.3	&18.3	&51.4 &\multirow{2}{*}{\textbf{1.8}} \\
\textbf{Ours (with SIM~\cite{SIM})} & ResNet101 &86.7 	&\textbf{81.9}	 &15.7	 &\textbf{17.7}	 &\textbf{31.7}	 &\textbf{82.3}	 &\textbf{48.2}	 &\textbf{86.6}	 &81.9	 &\textbf{32.3}	 &\textbf{20.4}	 &\textbf{53.2} &
\\
\bottomrule
\end{tabular}
}
\end{table*}

\begin{table}{}
\caption{Synthetic-to-real adaptation of object detection from SIM10K to Cityscapes (default dimension is style). 
}
\label{table:sim10k_det}
\centering
\resizebox{0.42\textwidth}{!}{%
\vspace{0.1cm}
\begin{tabular}{c|c|c|c}
\toprule
Method & backbone & Car AP & Gain\\ 
\midrule
Source Only~\cite{ren2016faster}  &ResNet50 & 42.8 & - \\
DA-Faster~\cite{DA-Faster-RCNN} & ResNet50 & 41.9 & -0.9\\
DD-MRL~\cite{DDMRL} & ResNet50 & 43.9  &1.1\\
SWDA~\cite{SCDA} & ResNet50 & 44.6  & 1.8\\
SCDA~\cite{SCDA} & ResNet50 & 45.1  & 2.3\\
MTOR~\cite{MTOR} & ResNet50 & 46.6 & 3.8\\
\midrule
GPA~\cite{GPA} & ResNet50 & 47.6 & 4.8\\
\textbf{Ours (with GPA)} & ResNet50 & \textbf{49.2} & \textbf{6.4}\\
\bottomrule
\end{tabular}
}
\end{table}

\paragraph{Cityscapes $\rightarrow$ RTTS.} RTTS~\cite{RTTS}
is the largest available dataset for object detection under real-world hazy conditions. It contains 4,807 unannotated and 4,322 annotated real-world hazy images covering 
most traffic and driving scenarios with 7 kinds of fogs. 

\paragraph{Cityscapes $\rightarrow$ Foggy Zurich++.} Foggy Zurich++ is a real-world foggy-weather dataset  for segmentation. We use all the unannotated 3,768 images of Foggy Zurich~\cite{FoggyCity} as the training set and mix the validation set of Foggy Driving~\cite{FoggyDriving} and Foggy Zurich~\cite{FoggyCity}. Following \cite{cordts2016cityscapes}, it is labeled with 19 classes. 

\paragraph{Cityscapes $\rightarrow$ RainCityscapes.} RainCityscapes~\cite{rainy_city} renders Cityscapes images with synthetic rain. Each clear image is rendered with 12 types of rain patterns, including 4 types of drop sizes. The annotations are the same as those of Cityscapes. We use this benchmark in cross-domain object detection.

\paragraph{VKITTI $\rightarrow$ CKITTI.}  We use this benchmark in both detection and segmentation. Virtual KITTI~\cite{VKITTI} is a photo-realistic synthetic dataset, which contains 21,260 images. 
It is designed to mimic the conditions of  KITTI dataset and 
has similar scene layouts, camera viewpoints and image resolution to KITTI dataset.
CKITTI is a real-world dataset depicting several urban driving scenarios with 5 different kinds of FoVs, which is a mixed dataset of Cityscapes~\cite{cordts2016cityscapes} and KITTI~\cite{kITTI}. We use the 10,456 images as the training set and 700 images as the validation set.

\paragraph{SIM10K $\rightarrow$ Cityscapes.} SIM10K~\cite{sim10k} contains 10,000 images of the computer-rendered driving scene from the Grand Theft Auto (GTAV) game. In this benchmark of object detection, SIM10K is the source domain and Cityscapes~\cite{cordts2016cityscapes} serves as the target domain.

\subsection{Implementation Details}
\label{sec:implementation details}
\paragraph{Object detection.}
In our implementation, we strictly follow the common training protocols~\cite{DA-Faster-RCNN,SWDA,SCL,GPA} of the Faster-RCNN network~\cite{ren2016faster}. We resize the images of both the source and target domains to 600-pixel height in all experiments as suggested by \cite{DA-Faster-RCNN,SWDA,SCL}. Following the aforementioned papers, we use VGG16~\cite{vgg} and ResNet50~\cite{he2016deep} pre-trained on ImageNet \cite{deng2009imagenet} as the backbone of DA-Faster~\cite{DA-Faster-RCNN}, SWDA~\cite{SWDA}, SCL~\cite{SCL} and GPA~\cite{GPA} for fair comparisons. We set the learning rate to 0.001 for the first 50k iterations and 0.0001 for the remaining iterations.  As suggested by the original authors~\cite{DA-Faster-RCNN,SWDA,SCL}, $\lambda_{adv}$ is set to 1.0, 1.0, 0.1 for ~\cite{SCL,SWDA,DA-Faster-RCNN}, respectively.
The IoU threshold 0.5 is used for evaluation and the mean average precision (mAP) is calculated as the evaluation metric.

\paragraph{Semantic segmentation.}
Following UDA protocols~\cite{AdaptSegNet,CLAN,SIM}, we employ the DeepLab-v2~\cite{chen2017deeplab} with ResNet 101 backbone~\cite{he2016deep} in our implementation. The backbone is pre-trained on ImageNet~\cite{deng2009imagenet}. We reproduce the famous  AdaptSegNet~\cite{AdaptSegNet}, CLAN~\cite{CLAN,CLANv2} and SIM~\cite{SIM} as our baselines. For our DeepLab-v2 network, we use Adam as the optimizer. The initial learning rate is  $2.5 \times 10 ^{-4}$, which is then decreased using polynomial decay with an exponent of $0.9$. We used the broadly utilized protocols,
per-class intersection-over-union (IoU) and mean IoU over
all categories for evaluation. As suggested by the original works, $\lambda_{adv}$ is set to 0.01, 0.001, 0.001 for \cite{AdaptSegNet,CLAN,SIM}, respectively.

\subsection{Domain Adaptation for Object Detection}
\label{sec:DAOD}
In this section, we perform cross-domain detection in four scenarios, \emph{i.e.,} cross-fog, cross-rain and cross-FoV adaptation, and  synthetic-to-real adaptation, to show the effectiveness of our approach. 

\begin{table}{}
\caption{Cross-fog adaptation of semantic segmentation from Cityscapes to Foggy Zurich++ (multi-domainness). }
\label{table:c2f_seg}
\centering
\resizebox{0.48\textwidth}{!}{%
\vspace{0.1cm}
\begin{tabular}{c|c|c|c}
\toprule
Method &backbone & mIoU
& Gain\\ 
\toprule
AdaptSegNet~\cite{AdaptSegNet}  &ResNet101 &29.4 & \multirow{2}{*}{\textbf{5.8}}\\
\textbf{Ours (with AdaptSegNet~\cite{AdaptSegNet})} &ResNet101 &\textbf{35.2} &\\
\midrule
CLAN~\cite{CLAN} &ResNet101 &26.8 &	\multirow{2}{*}{\textbf{4.7}}\\
\textbf{Ours (with CLAN~\cite{CLAN})} &ResNet101	&\textbf{31.5}	&\\
\midrule
SIM~\cite{SIM} &ResNet101	&27.0	&\multirow{2}{*}{\textbf{4.1}} \\
\textbf{Ours (with SIM~\cite{SIM})} &ResNet101	&\textbf{31.1}	&\\
\bottomrule
\end{tabular}
}
\end{table}

\paragraph{Cross-fog adaptation.}
To validate the generalization capability on the cross-fog adaptation, we perform two experiments, where the target domain includes single and multiple domainness values.

\noindent  
\emph{Single domainness within the target domain:}
Table \ref{table:ob_comb}~(a) presents the comparison results with the state-of-the-art cross-domain detection methods with VGG16~\cite{vgg} and ResNet50~\cite{he2016deep}. In this experiment, we adapt from Cityscapes~\cite{cordts2016cityscapes} to Foggy-Cityscapes~\cite{FoggyCity}. Source-only indicates the baseline Faster RCNN~\cite{ren2015faster} is trained with the source domain only.  With VGG16~\cite{vgg} as the backbone, the DA-Faster~\cite{DA-Faster-RCNN} and GPA~\cite{GPA} baselines are 27.6\% and 37.5\% mAP, respectively. By plugging into them, our proposed method is superior to these baselines by 3.6\% and 5.4\%, achieving 31.2\% and 42.9\% mAP, respectively, which demonstrates the effectiveness of our method. 
For the VGG16-based methods, our result outperforms the state-of-the-art methods
by at least 1.1\%. Compared to the ResNet50-based methods, we outperform all prior works and get a significant mAP gain of +5.7\% over the GPA~\cite{GPA} baseline, achieving 45.7\% mAP.  Consistent improvements with different backbones illustrate
the effectiveness of the proposed method.

Taking a closer look at per-category  performance in Table \ref{table:ob_comb}~(a), our approach achieves the highest AP on most categories.
This phenomenon illustrates the effectiveness of the proposed SAD among different classes during the adaptation. Interestingly, as shown in Table \ref{table:ob_comb}~(a), the result of our proposed method ($45.2\%$ mAP) exceeds the oracle result ($43.8\%$ mAP) on this dataset, showing that the diversified images generated by DC including clear weather images with high visibility and adverse weather images with low visibility are useful for boosting the adaptation, which also indicates that learning domainness-invariant features is beneficial to bridge the intra-domain gap.

\noindent \emph{Multiple domainness within the target domain:} In this experiment, we adapt from Cityscapes~\cite{cordts2016cityscapes} to RTTS dataset~\cite{RTTS}. Multi-domainness means there exist 7 kinds of fogs in RTTS dataset.
The comparison results with the state-of-the-arts are reported in Table \ref{table:ob_comb}~(b). As for the image dehazing approaches which dehaze the target domain and then trasfer the domain knowledge, DCPDN~\cite{zhang2018densely} improves the Faster RCNN performance by $0.7\%$. However, Grid-Dehaze~\cite{liu2019griddehazenet} does not help the Faster RCNN baseline and results in even worse performance. 
Table \ref{table:ob_comb}~(b) shows that our method can effectively boost the performance by integrating it into   DA-Faster RCNN~\cite{DA-Faster-RCNN} and SWDA~\cite{SWDA}. We successfully boost the mAP by $2.0\%$ and $2.0\%$, respectively. The benefits of our approach lie in two aspects: (1) our method can be easily adopted as a plug-and-play framework during the training and does not introduce any extra costs in the inference time. (2) Our approach not only address the single domainness problem but also tackle more complicated scenarios where multiple domainness exist in the target domain.

\paragraph{Cross-FoV adaptation.} To validate the generalization capability of the proposed method, we also conduct an experiment on the FoV dimension adapting from Virtual KITTI~\cite{VKITTI} to CKITTI~\cite{cordts2016cityscapes,kITTI}. 
The adaptation results are reported in Table \ref{table:ob_comb}~(c). Despite the 5 different FoVs in the dataset, our method always achieves consistent improvements. By plugging into the current state-of-the-art methods, \emph{i.e.,} DA-Faster~\cite{DA-Faster-RCNN}, SWDA~\cite{SWDA}, SCL~\cite{SCL},
our method brings $2.6\%$, $1.7\%$ and $1.8\%$ increase, respectively.

\paragraph{Cross-rain adaptation.} 
Table \ref{table:ob_comb} (d) shows the adaptation results between different rain scenarios on Cityscapes~\cite{cordts2016cityscapes} to RainCityscapes~\cite{rainy_city}. We reproduce DA-Faster RCNN~\cite{DA-Faster-RCNN},  SWDA~\cite{SWDA} and SCL~\cite{SCL} in the same setting. From the table, we can observe that our method significantly improves the mAP by $6.4\%$, $3.4\%$, and $4.2\%$, respectively, by integrating it into the existing UDA methods.

\paragraph{Synthetic-to-real adaptation.}
In this setting, we adapt from SIM10K~\cite{sim10k} to Cityscapes~\cite{cordts2016cityscapes} to study the scenario in which we do not have any prior knowledge about the domain shifts. Under such cases, our DC will utilize the style of the target domain by default as prior knowledge. As shown in Table~\ref{table:sim10k_det}, our approach still achieves obvious improvements over the strong baseline~\cite{GPA}, which achieves a new state-of-the-art $49.2\%$ AP. 
The main reason is that our DC generates diversified images with different domainness of styles, and our method can well learn texture-invariant features to bridge the domain gap.

\subsection{Domain Adaptation for Semantic Segmentation}
\label{sec:DASS}
In addition to cross-domain object detection, we also conduct experiments on cross-domain semantic segmentation, to show the scalability of our method. In particular, we conduct the cross-FoV adaptation and cross-fog adaptation on semantic segmentation.

\paragraph{Cross-fog adaptation.}
In this experiment, we adapt from Cityscapes~\cite{cordts2016cityscapes} to Foggy Zurich++~\cite{FoggyCity,FoggyDriving} to perform the cross-fog adaptation, where multiple degrees of fog thickness exist in the target domain. As shown in Table \ref{table:c2f_seg}, our method outperforms the state-of-the-art methods~\cite{AdaptSegNet,CLAN,SIM} by $5.8\%$, $4.7\%$ and $4.1\%$, respectively. 
Our method can handle the cases where a domainness value is never seen in the training stage. As shown in Table \ref{table:c2f_seg}, the Foggy Zurich++ has the real fog rather than the synthetic fog, which means the domainness in the validation set is unknown and does not appear in the training set. Our method works well on this dataset, which proves its generalization ability. 

\paragraph{Cross-FoV adaptation.}
In this experiment, we perform the specific domain adaptation given the FoV gap. We choose Virtual KITTI~\cite{VKITTI} as the source domain and CKITTI~\cite{kITTI,cordts2016cityscapes} as the target domain. The comparison results are listed in Table \ref{table:v2c_seg}. Compared with the AdaptSegNet~\cite{AdaptSegNet}, CLAN~\cite{CLAN} and SIM~\cite{SIM}, our method respectively yields an increase of $1.8\%$, $1.1\%$ and $1.8\%$, which indicates the effectiveness of the proposed SAD in the semantic segmentation task and shows its good scalability.

\subsection{Ablation Studies and Analysis}
\label{sec:ablation}
In this section, we perform thorough ablation studies with visualizations to investigate the effect of each component. 
\subsubsection{Ablation Studies of Components}

\noindent \textbf{Effects of each component on different tasks.} 
Table \ref{table:ablation3} summarizes the effects of each designed component on cross-domain object detection and semantic segmentation. The former is conducted on  Cityscapes~\cite{cordts2016cityscapes} $\rightarrow$  Foggy Cityscapes~\cite{FoggyCity}. The latter is adapted from Cityscapes~\cite{cordts2016cityscapes} to Foggy Zurich++~\cite{FoggyCity,FoggyDriving}. 
The GPA~\cite{GPA} baseline is $39.5\%$.  
By adding the DC and SAR sequentially, we boost the mAP with an additional $+3.0\%$ and $+2.7\%$, achieving $42.5\%$ and $45.2\%$, respectively. Similarly, we boost the AdaptSegNet~\cite{AdaptSegNet} baseline by $+4.1\%$ and $+1.7\%$, achieving $33.5\%$ and $35.2\%$, respectively.
These improvements in two tasks reveal the effects of individual components of our approach. It also shows that these two components are complementary and together they significantly promote the performance.

\begin{table}[t]
\caption{
Ablation study of each component on two tasks.
}
\label{table:ablation3}
\centering
\resizebox{0.4\textwidth}{!}{%
\begin{tabular}{ccc|c|c} \toprule
GPA~\cite{GPA} & DC & SAR  & mAP & Gain\\
\midrule
$\surd$  &  &  & 39.5 & -\\
$\surd$  & $\surd$  &  & 42.5 & 3.0\\
$\surd$  & $\surd$  & $\surd$  & 45.2 & 5.7\\
\midrule
AdaptSegNet~\cite{AdaptSegNet} & DC & SAR  & mIoU & Gain\\
\midrule
$\surd$  &  &  & 29.4 & -\\
$\surd$  & $\surd$  &  & 33.5 & 4.1\\
$\surd$  & $\surd$  & $\surd$  & 35.2 & 5.8\\
\bottomrule
\end{tabular}
}
\end{table}

\noindent \textbf{Comparisons to random data augmentation with labels.}  
As shown in Table~\ref{table:comparison_augmentations}, we compared our DC with two representative data augmentations, \emph{i.e.,} random crop, color jittering, and labels of the augmentations are also used to guide the SAR. Random crop and color jittering achieve less substantial improvements over the strong baseline from Cityscapes~\cite{cordts2016cityscapes} to Foggy Cityscapes~\cite{FoggyCity}, reaching $42.7\%$ and $41.8\%$ mAP, respectively. The main reason for improvements is the fact that FoV shifts and color shifts are generally the subsets of domain shifts, and such shifts indeed improve the performance though they are not the main dimension of domain shifts. It also demonstrates that the proposed SAR module is general and effective with regard to different dimensions of domainness. In contrast to these random augmentations, by modeling the specific domain dimension in DC, our proposed method achieves a much better performance ($45.2\%$ mAP), which soundly confirms the motivation that such prior knowledge of the target domain is useful for adapting to a specific domain.

\begin{table}[t]
\caption{
Comparison to data augmentations with labels. 
}
\centering
\resizebox{0.40\textwidth}{!}{%
\label{table:comparison_augmentations}
\begin{tabular}{c|c|c} \toprule
Ablations & mAP & Gain\\
\midrule
GPA Baseline~\cite{GPA} & 39.5 & -\\
Ours (w/o DC) + Color Jittering & 41.8 & 2.3 \\
Ours (w/o DC) + Random Crop & 42.7 & 3.2\\
Ours (w DC) & 45.2 & 5.7\\
\bottomrule
\end{tabular}
}
\end{table}

\begin{table}[t]
\caption{ Ablation study of the losses $\mathcal{L}_{inv}$ and $\mathcal{L}_{spf}$.
}
\label{table:ablation_each_loss}
\begin{center}
\vspace{-3mm}
\begin{tabular}{ccc|c|c} \toprule
GPA~\cite{GPA} + DC & $\mathcal{L}_{inv}$ & $\mathcal{L}_{spf}$  & mAP & Gain\\
\midrule
$\surd$  &  &  & 42.5 & -\\
$\surd$  & $\surd$  &  & 43.3 & 0.8\\
$\surd$  &   & $\surd$  & 43.1 & 0.6\\
$\surd$  & $\surd$  & $\surd$  & 45.2 & 2.7\\
\midrule
AdaptSegNet~\cite{AdaptSegNet} + DC & $\mathcal{L}_{inv}$  & $\mathcal{L}_{spf}$   & mIoU & Gain\\
\midrule
$\surd$  &  &  & 33.5 & -\\
$\surd$  & $\surd$  &  & 34.1 & 0.6\\
$\surd$  &  & $\surd$  & 34.0 & 0.5\\
$\surd$  & $\surd$ & $\surd$  & 35.2 & 1.7\\
\bottomrule
\end{tabular}
\end{center}
\end{table}

\begin{table}[t]
\caption{
Comparison to related works on the baseline~\cite{AdaptSegNet}. 
}

\centering
\resizebox{0.40\textwidth}{!}{%
\label{table:comparison_related_works}

\begin{tabular}{c|c|c} \toprule
Ablations & mIoU & Gain\\
\midrule
Ours (w/o SAR) & 33.5 & - \\
Ours (w/o SAR) + MRL~\cite{DDMRL} & 34.0 & 0.5\\
Ours (w/o SAR) + CIDA~\cite{CIDA} & 34.2 & 0.7\\
Ours (w SAR) & 35.2 & 1.7\\
\bottomrule
\end{tabular}
}
\end{table}

\noindent \textbf{Ablation of loss functions on different tasks.} 
Table~\ref{table:ablation_each_loss}  shows the ablation of the domainnness-specific loss $\mathcal{L}_{spf}$ (Eq.(\ref{eq:domainness-specific})) and domainness-invariant loss $\mathcal{L}_{spf}$ (Eq.(\ref{eq:domainness-inv})) in two tasks. 
GPA~\cite{GPA} is the base network for object detection adapting from Cityscapes~\cite{cordts2016cityscapes} $\rightarrow$  Foggy Cityscapes~\cite{FoggyCity}. Besides, for semantic segmentation, our method is adapted from  Cityscapes~\cite{cordts2016cityscapes} to Foggy Zurich++~\cite{FoggyCity,FoggyDriving} with AdaptSegNet~\cite{AdaptSegNet}. 
As shown in Table~\ref{table:ablation_each_loss}, we observe that merely using domainnness-specific loss $\mathcal{L}_{spf}$ or domainness-invariant loss $\mathcal{L}_{inv}$ cannot achieve huge improvements over the baseline. The main reason is that domainnness-invariant loss $\mathcal{L}_{inv}$ and domainnness-specific loss $\mathcal{L}_{spf}$ are both critical for learning the domainnesss-invariant representations for the self-adversarial disentangling, and together they promote the disentangling in opposite directions, \emph{i.e.,} a self-adversarial manner. 
This shows that our SAR needs to be trained under the guidance of both loss functions,  \emph{i.e.,} $\mathcal{L}_{spf}$ and $\mathcal{L}_{inv}$. 

\noindent \textbf{Comparisons to the related works.}
Table \ref{table:comparison_related_works} shows the comparisons to the relevant works~\cite{DDMRL, CIDA} from Cityscapes~\cite{cordts2016cityscapes} to Foggy Zurich++~\cite{FoggyCity,FoggyDriving} under the same baseline~\cite{AdaptSegNet}.
As we can see, when using MRL~\cite{DDMRL} or CIDA~\cite{CIDA} as the adaptor, it merely achieves a limited improvement of $0.5\%$ or $0.7\%$. In contrast, SAR  contributes to the performance gain of $1.7\%$. 
The main reasons are twofold. (1) Previous GAN-based methods~\cite{DDMRL, CIDA} do not utilize supervisory signals $d'_{gt}$ from DC to fully learn the domainness-invariant feature. (2) They neglect the intra-domain gap induced by different domainness. Instead, our method not only leverages the prior supervisory signals but also mitigates the intra-domain gap across different domainness. Incorporating DC and SAR into the same framework boosts the mIoU by $1.7\%$ over the baseline. This confirms the effectiveness of our proposed DC and SAR, and addresses the aforementioned claim in Section \ref{sec: sar} that our SAD framework is superior to GAN.

\begin{figure*}[t]
\centering
\includegraphics[scale=0.9]{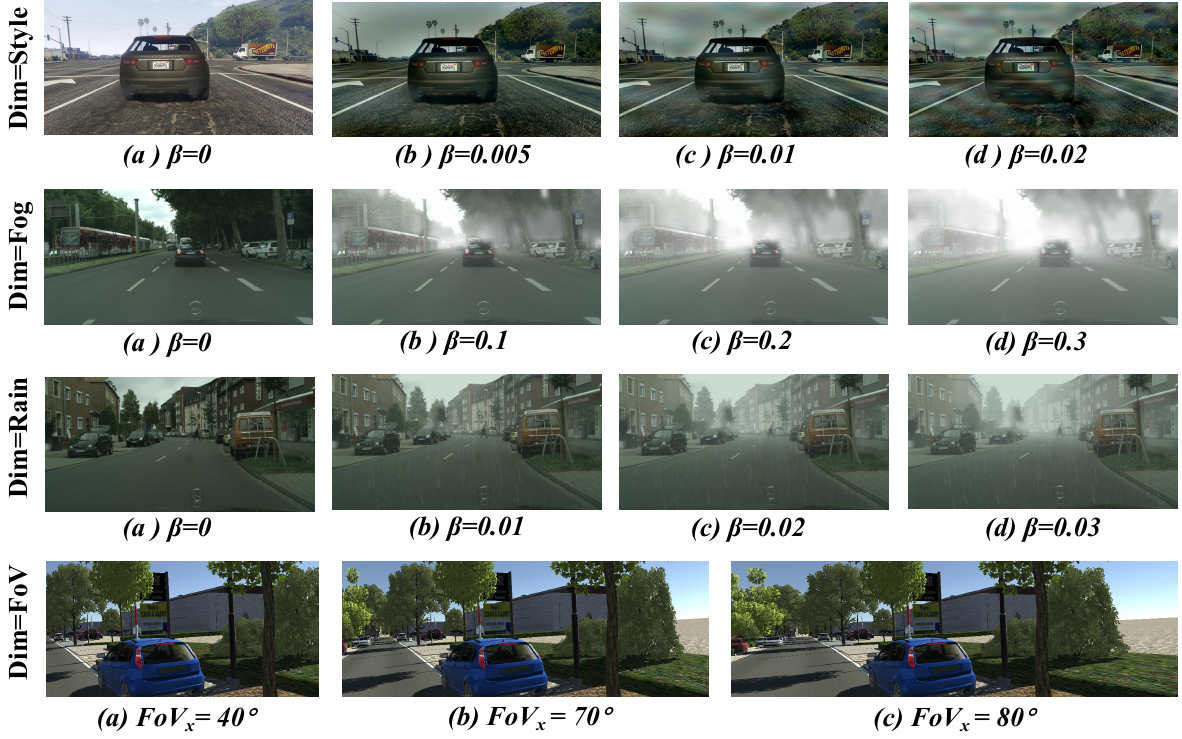}
\caption{
Visualization of diversified source images produced by Domainness Creator with different domainness values on style, fog, rain and FoV dimensions, respectively. 
With the increased variations of domainness, our model can learn the domainness-invariant features (Best viewed in color).
}
\label{fig:dc_results}
\end{figure*}
\subsubsection{Hyper-parameter Analysis}

\begin{figure}[t]{
\centering
    \includegraphics[scale=0.31]{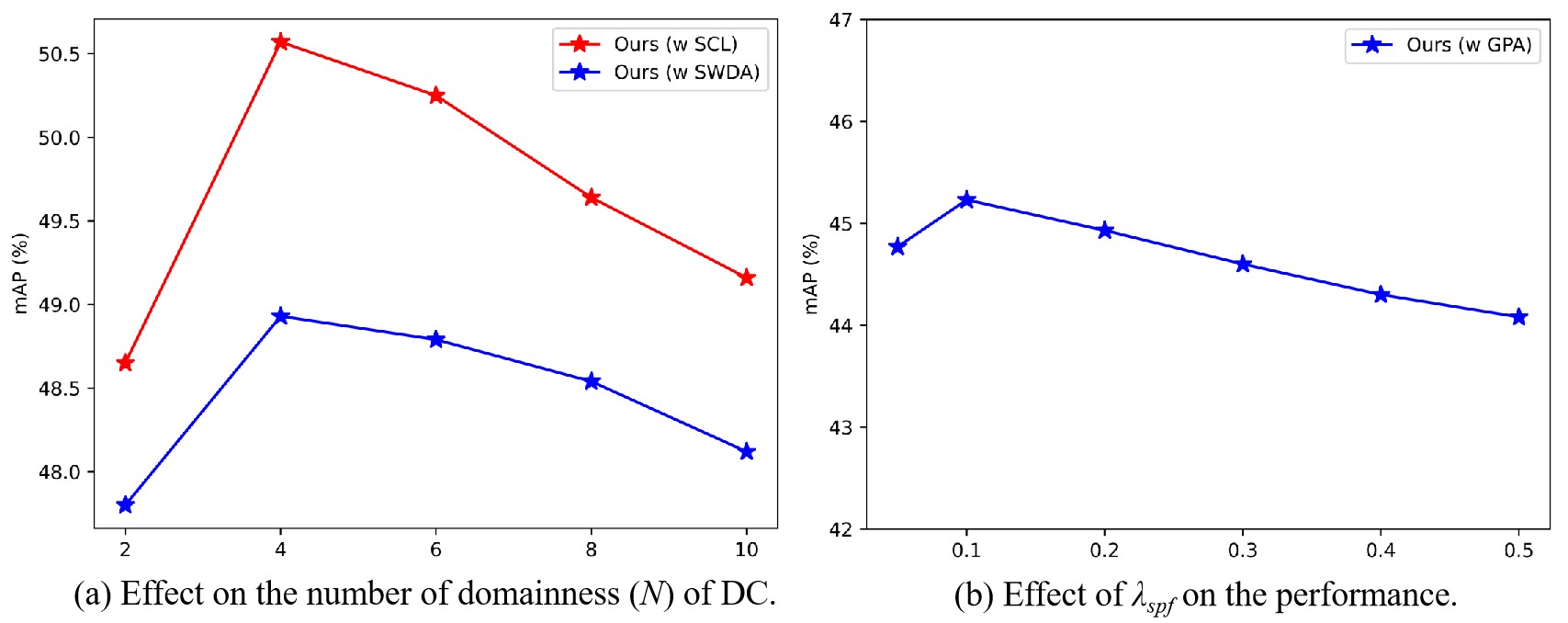}
    \captionof{figure}{
    Hyper-parameter analysis: (a) effect of domainness number ($N$) of DC with SCL~\cite{SCL} and SWDA~\cite{SWDA} on VKIITI~\cite{VKITTI} to CKITTI~\cite{kITTI,cordts2016cityscapes}. (b) Effect of $\lambda_{spf}$ on the performance with GPA~\cite{GPA} on Cityscapes~\cite{cordts2016cityscapes} to Foggy Cityscapes~\cite{FoggyCity} (Best viewed in color).
    }
\label{fig:vis_params}}
\vspace{-3mm}
\end{figure}

\noindent \textbf{Effect of the domainness number of DC.}  
During the adaptation, we do not pre-know the existing number of domainness values ($N$) in the given target dataset, and thus we conduct ablation experiments on different adaptation benchmarks to study the effects of the number of domainness values ($N$) on the final performance. 
As shown in Fig.~\ref{fig:vis_params} (a), we plot the performance of the model trained with different domainness numbers ($N$) in the cross-FoV adaptation benchmark (Virtual KITTI~\cite{VKITTI} to CKITTI~\cite{cordts2016cityscapes,kITTI})  with different baseline models (SCL~\cite{SCL} and SWDA~\cite{SWDA}). As we can see, when $N$ is too small and too large, the performances are less desired. We observe that the best performance occurs when $N$ is around 4. The main reasons behind this phenomenon can be explained as follows. The too-fine division will make the differences between two identical classes with different domainness values small, while too-coarse divisions will be insufficient to constrain the disentangling. After being divided into this number of domnainness, DC could produce transformed images that have obvious visual differences with different domainness values, which is more suitable for disentangling. Thus, such a number of domainness will be a better choice during the adaptation without the prior knowledge $N$ of the target domain. 
To confirm whether such a choice performs well in other adaptation benchmarks, we conduct more experiments on different dimensions to study the effect of $N$ on the final performance. As shown in Table \ref{tabble:ablation_domainness}, by gradually increasing $N$ to 4, we observe that the performance also improves, which confirms the aforementioned claims. To show the robustness of the proposed method, we set $N=4$ in all experiments. 

\begin{table}[t!]
    \footnotesize
    \begin{center}
    \vspace{-2mm}
    \addtolength{\tabcolsep}{1.5pt}
    \caption{
    Ablations on the domainness number ($N$) of DC.
    }
\resizebox{0.43\textwidth}{!}{%
    \label{tabble:ablation_domainness}
    \begin{tabular}{c|ccccc}
\Xhline{1.0pt}
\multicolumn{6}{c}{(a) {Number of domainness on \textit{style} (default) dimension.}}\\
\hline
$N$ &  0 & 1 & 2 & 3 & 4\\
mAP(\%) & 48.52 & 48.74 & 48.77 & 48.93 & 49.17 \\
\Xhline{1.0pt}
\multicolumn{6}{c}{(b) {Number of domainness on \textit{fog} dimension.}}\\
\hline
$N$ & 0 & 1 & 2 & 3  & 4  \\
mAP(\%)  & 43.17  & 43.47  & 43.52 & 43.94  & 45.23 \\
\Xhline{1.0pt}
\multicolumn{6}{c}{(c) {Number of domainness on \textit{rain} dimension.}}\\
\hline
$N$ & 0 & 1 & 2  & 3 & 4  \\
mAP (\%)  & 34.96 & 41.23 & 41.72 & 42.67 & 44.21 \\
\Xhline{1.0pt}
\multicolumn{6}{c}{(d) {Number of domainness on \textit{FoV} dimension.}}\\
\hline
$N$ & 0 & 1 & 2  & 3 & 4\\
mAP(\%) &46.14  & 47.00 & 48.57 & 48.57 &  50.57 \\
\Xhline{1.0pt}
    \end{tabular}
    }
    \end{center}
\end{table}

\begin{figure*}[htbp]
\centering
\includegraphics[scale=1.0]{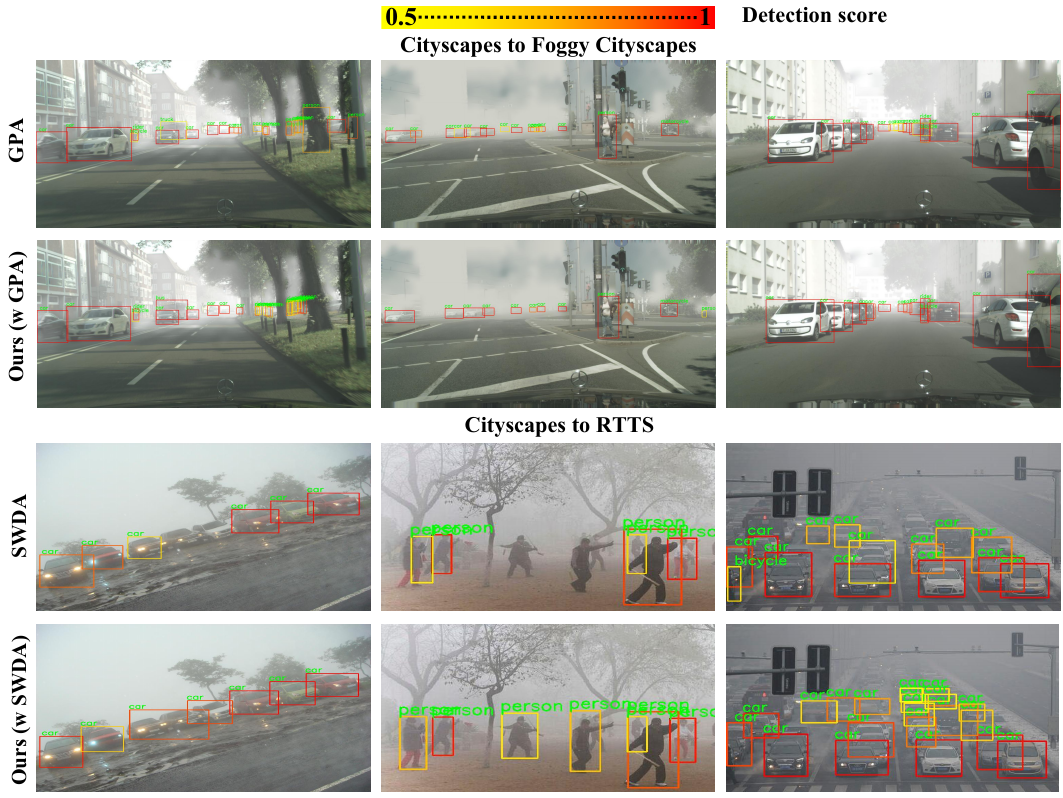}
\caption{
Qualitative results of cross-domain object detection on Cityscapes~\cite{cordts2016cityscapes} $\rightarrow$ Foggy Cityscapes~\cite{FoggyCity} and Cityscapes~\cite{cordts2016cityscapes} $\rightarrow$ RTTS~\cite{RTTS} set-up. The first and third rows plot the predictions of GPA~\cite{GPA} and SWDA~\cite{SWDA} baseline, and the second and fourth rows plot the predictions of Ours (with GPA~\cite{GPA} and SWDA~\cite{SWDA}). Bounding boxes are colored based on the detector's confidence using the shown color map. 
Our method could detect more objects in the images accurately.
}
\label{fig:c2f_det}
\vspace{-1mm}
\end{figure*}

 \begin{figure*}[t!]{
\centering
    \includegraphics[scale=1.05]{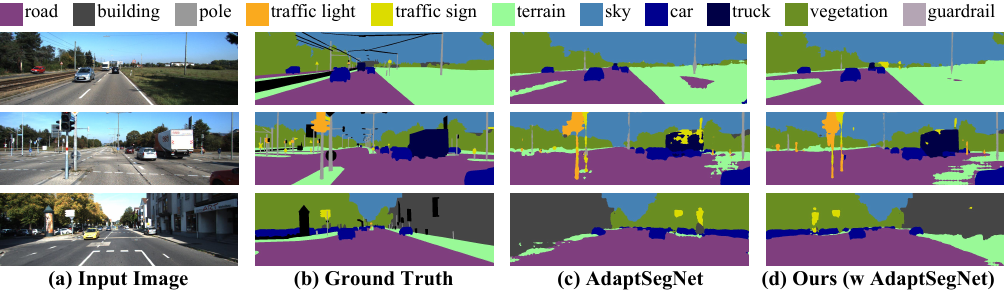}
    \captionof{figure}{
    Qualitative results of cross-domain semantic segmentation on Virtual KITTI~\cite{VKITTI} $\rightarrow$  CKITTI~\cite{kITTI,cordts2016cityscapes} set-up. The four columns plot (a) input RGB  images, (b) ground truth, (c) the predictions of AdaptsegNet~\cite{AdaptSegNet} baseline, and (d) Ours (with AdaptsegNet~\cite{AdaptSegNet}).  Our method could segment more precisely than prior work~\cite{AdaptSegNet}. (Best viewed in color). 
    }
\label{fig:v2c_kitti}}
\vspace{-3mm}
\end{figure*}

\noindent \textbf{Parameter Analysis of $\lambda_{spf}$.}  
We investigate the effect of the hyper-parameter $\lambda_{spf}$ which balances the domain adaptation process. 
In Fig.~\ref{fig:vis_params} (b), we plot the performance curve of models trained with different $\lambda_{spf}$ values on the setting of Cityscapes~\cite{cordts2016cityscapes} $\rightarrow$ Foggy Cityscapes~\cite{FoggyCity} in object detection task.  
The highest mAP on the target domain is
achieved when the value of  $\lambda_{spf}$  is around $0.1$, which means that this weight among
different loss functions benefits domain adaptation the most. 
We simply set the same $\lambda_{spf}=0.1$ in all experiments to show the robustness of our method in different settings.  

\subsubsection{Visualization of Predictions and Generated Results}
\label{sec:visualization}
Fig.~\ref{fig:dc_results} shows the visualization of diversified source images produced by our Domainness Creator with different domainness values on style, fog, rain and FoV dimensions, respectively. For example, as shown in the last row, from left to right it displays the processed image with $FoV_{x}$ of $40^{\circ}$,  $70^{\circ}$ and $80^{\circ}$, respectively. Due to the increased variations of domainness, a model  trained on this domainness-diversified dataset is able to learn the domainness-invariant representation for specific domain adaptation. 
 
Fig.~\ref{fig:c2f_det} visualizes the qualitative results of cross-domain object detection on two benchmarks, Cityscapes~\cite{cordts2016cityscapes} $\rightarrow$ Foggy Cityscapes~\cite{FoggyCity} and Cityscapes~\cite{cordts2016cityscapes} $\rightarrow$ RTTS~\cite{RTTS}, respectively. As we can see, our method is able to produce high confidence predictions and is able to detect more objects when plugging into the current state-of-the-art methods, \emph{e.g.,} GPA and SWDA~\cite{SWDA}.

Fig.~\ref{fig:v2c_kitti} shows the qualitative results of cross-domain semantic segmentation from Virtual KITTI dataset~\cite{VKITTI} to CKITTI~\cite{kITTI,cordts2016cityscapes}. With the aid of our proposed Self-Adversarial Disentangling framework,  our models (with AdaptSegNet~\cite{AdaptSegNet}) are able to produce correct predictions at a high level of confidence, and the proposed method achieves good performance on most categories, \emph{e.g.,} ‘vegetation’, ‘terrain’, ‘car’, 'truck', and ‘traffic sign’ classes. 

\section{Conclusion}
In this paper, we studied specific domain adaptation (SDA) and proposed self-adversarial disentangling (SAD) to learn domainness-invariant features in a specific dimension. The domainness creator aims to enrich the source domain and to provide additional supervisory signals for fully learning the domainness-invariant feature. The self-adversarial regularizer and two losses are introduced to narrow the intra-domain gap induced by different domainness. Extensive experiments validate our method on object detection and semantic segmentation under various domain-shift settings. Our method can be easily integrated into state-of-the-art architectures to attain considerable performance gains. 

\section*{Acknowledgment}
The authors would like to thank Zhengyang Feng (Shanghai Jiao Tong University), Guangliang Cheng and Jianping Shi (Sensetime) for their suggestions.


\ifCLASSOPTIONcaptionsoff
  \newpage
\fi



%
\bibliographystyle{IEEEtran}
\bibliography{egbib}

\begin{thebibliography}{10}
\providecommand{\url}[1]{#1}
\csname url@samestyle\endcsname
\providecommand{\newblock}{\relax}
\providecommand{\bibinfo}[2]{#2}
\providecommand{\BIBentrySTDinterwordspacing}{\spaceskip=0pt\relax}
\providecommand{\BIBentryALTinterwordstretchfactor}{4}
\providecommand{\BIBentryALTinterwordspacing}{\spaceskip=\fontdimen2\font plus
\BIBentryALTinterwordstretchfactor\fontdimen3\font minus
  \fontdimen4\font\relax}
\providecommand{\BIBforeignlanguage}[2]{{%
\expandafter\ifx\csname l@#1\endcsname\relax
\typeout{** WARNING: IEEEtran.bst: No hyphenation pattern has been}%
\typeout{** loaded for the language `#1'. Using the pattern for}%
\typeout{** the default language instead.}%
\else
\language=\csname l@#1\endcsname
\fi
#2}}
\providecommand{\BIBdecl}{\relax}
\BIBdecl

\bibitem{girshick2014rich}
R.~Girshick, J.~Donahue, T.~Darrell, and J.~Malik, ``Rich feature hierarchies
  for accurate object detection and semantic segmentation,'' in
  \emph{Proceedings of the IEEE conference on computer vision and pattern
  recognition}, 2014, pp. 580--587.

\bibitem{ren2016faster}
S.~Ren, K.~He, R.~Girshick, and J.~Sun, ``Faster r-cnn: towards real-time
  object detection with region proposal networks,'' \emph{IEEE transactions on
  pattern analysis and machine intelligence}, vol.~39, no.~6, pp. 1137--1149,
  2016.

\bibitem{chen2017s}
T.~Chen, S.~Lu, and J.~Fan, ``S-cnn: Subcategory-aware convolutional networks
  for object detection,'' \emph{IEEE transactions on pattern analysis and
  machine intelligence}, vol.~40, no.~10, pp. 2522--2528, 2017.

\bibitem{focal_loss}
T.-Y. Lin, P.~Goyal, R.~Girshick, K.~He, and P.~Dollár, ``Focal loss for dense
  object detection,'' \emph{IEEE Transactions on Pattern Analysis and Machine
  Intelligence}, vol.~42, no.~2, pp. 318--327, 2020.

\bibitem{shen2019object}
Z.~Shen, Z.~Liu, J.~Li, Y.-G. Jiang, Y.~Chen, and X.~Xue, ``Object detection
  from scratch with deep supervision,'' \emph{IEEE transactions on pattern
  analysis and machine intelligence}, vol.~42, no.~2, pp. 398--412, 2019.

\bibitem{cai2019cascade}
Z.~Cai and N.~Vasconcelos, ``Cascade r-cnn: High quality object detection and
  instance segmentation,'' \emph{IEEE Transactions on Pattern Analysis and
  Machine Intelligence}, 2019.

\bibitem{oksuz2020imbalance}
K.~Oksuz, B.~C. Cam, S.~Kalkan, and E.~Akbas, ``Imbalance problems in object
  detection: A review,'' \emph{IEEE transactions on pattern analysis and
  machine intelligence}, 2020.

\bibitem{Tan_2022_TPAMI_mirror}
X.~Tan, J.~Lin, K.~Xu, C.~Pan, L.~Ma, and R.~W.~H. Lau, ``Mirror detection with
  the visual chirality cue,'' \emph{IEEE Transactions on Pattern Analysis and
  Machine Intelligence}, 2022.

\bibitem{zhou222transvod}
Q.~Zhou, X.~Li, L.~He, Y.~Yang, G.~Cheng, Y.~Tong, L.~Ma, and D.~Tao,
  ``Transvod: End-to-end video object detection with spatial-temporal
  transformers,'' \emph{IEEE Transactions on Pattern Analysis and Machine
  Intelligence}, pp. 1--16, 2022.

\bibitem{fcn}
J.~Long, E.~Shelhamer, and T.~Darrell, ``Fully convolutional networks for
  semantic segmentation,'' in \emph{Proceedings of the IEEE conference on
  computer vision and pattern recognition}, 2015, pp. 3431--3440.

\bibitem{chen2017deeplab}
L.~Chen, G.~Papandreou, I.~Kokkinos, K.~Murphy, and A.~L. Yuille, ``Deeplab:
  Semantic image segmentation with deep convolutional nets, atrous convolution,
  and fully connected crfs,'' \emph{{IEEE} Trans. Pattern Anal. Mach. Intell.},
  vol.~40, no.~4, pp. 834--848, 2018.

\bibitem{chen2018encoder}
L.-C. Chen, Y.~Zhu, G.~Papandreou, F.~Schroff, and H.~Adam, ``Encoder-decoder
  with atrous separable convolution for semantic image segmentation,'' in
  \emph{Proceedings of the European conference on computer vision (ECCV)},
  2018, pp. 801--818.

\bibitem{zhao2017pspnet}
H.~Zhao, J.~Shi, X.~Qi, X.~Wang, and J.~Jia, ``Pyramid scene parsing network,''
  in \emph{Proceedings of the IEEE conference on computer vision and pattern
  recognition}, 2017, pp. 2881--2890.

\bibitem{feng2020dmt}
Z.~Feng, Q.~Zhou, Q.~Gu, X.~Tan, G.~Cheng, X.~Lu, J.~Shi, and L.~Ma, ``Dmt:
  Dynamic mutual training for semi-supervised learning,'' \emph{Pattern
  Recognition}, p. 108777, 2022.

\bibitem{lin2017exploring}
G.~Lin, C.~Shen, A.~Van Den~Hengel, and I.~Reid, ``Exploring context with deep
  structured models for semantic segmentation,'' \emph{IEEE transactions on
  pattern analysis and machine intelligence}, vol.~40, no.~6, pp. 1352--1366,
  2017.

\bibitem{badrinarayanan2017segnet}
V.~Badrinarayanan, A.~Kendall, and R.~Cipolla, ``Segnet: A deep convolutional
  encoder-decoder architecture for image segmentation,'' \emph{IEEE
  transactions on pattern analysis and machine intelligence}, vol.~39, no.~12,
  pp. 2481--2495, 2017.

\bibitem{night_city}
X.~Tan, K.~Xu, Y.~Cao, Y.~Zhang, L.~Ma, and R.~W. Lau, ``Night-time scene
  parsing with a large real dataset,'' \emph{IEEE Transactions on Image
  Processing}, vol.~30, pp. 9085--9098, 2021.

\bibitem{DA-Faster-RCNN}
Y.~Chen, W.~Li, C.~Sakaridis, D.~Dai, and L.~Van~Gool, ``Domain adaptive faster
  r-cnn for object detection in the wild,'' in \emph{Proceedings of the IEEE
  conference on computer vision and pattern recognition}, 2018, pp. 3339--3348.

\bibitem{SWDA}
K.~Saito, Y.~Ushiku, T.~Harada, and K.~Saenko, ``Strong-weak distribution
  alignment for adaptive object detection,'' in \emph{Proceedings of the
  IEEE/CVF Conference on Computer Vision and Pattern Recognition}, 2019, pp.
  6956--6965.

\bibitem{MAF}
Z.~He and L.~Zhang, ``Multi-adversarial faster-rcnn for unrestricted object
  detection,'' in \emph{Proceedings of the IEEE/CVF International Conference on
  Computer Vision}, 2019, pp. 6668--6677.

\bibitem{SCDA}
X.~Zhu, J.~Pang, C.~Yang, J.~Shi, and D.~Lin, ``Adapting object detectors via
  selective cross-domain alignment,'' in \emph{Proceedings of the IEEE/CVF
  Conference on Computer Vision and Pattern Recognition}, 2019, pp. 687--696.

\bibitem{GPA}
M.~Xu, H.~Wang, B.~Ni, Q.~Tian, and W.~Zhang, ``Cross-domain detection via
  graph-induced prototype alignment,'' in \emph{Proceedings of the IEEE/CVF
  Conference on Computer Vision and Pattern Recognition}, 2020, pp.
  12\,355--12\,364.

\bibitem{HTCN}
C.~Chen, Z.~Zheng, X.~Ding, Y.~Huang, and Q.~Dou, ``Harmonizing transferability
  and discriminability for adapting object detectors,'' in \emph{Proceedings of
  the IEEE/CVF Conference on Computer Vision and Pattern Recognition}, 2020,
  pp. 8869--8878.

\bibitem{ICR-CCR}
C.-D. Xu, X.-R. Zhao, X.~Jin, and X.-S. Wei, ``Exploring categorical
  regularization for domain adaptive object detection,'' in \emph{Proceedings
  of the IEEE/CVF Conference on Computer Vision and Pattern Recognition}, 2020,
  pp. 11\,724--11\,733.

\bibitem{SCL}
Z.~Shen, H.~Maheshwari, W.~Yao, and M.~Savvides, ``Scl: Towards accurate domain
  adaptive object detection via gradient detach based stacked complementary
  losses,'' \emph{arXiv preprint arXiv:1911.02559}, 2019.

\bibitem{hsu2020every}
C.-C. Hsu, Y.-H. Tsai, Y.-Y. Lin, and M.-H. Yang, ``Every pixel matters:
  Center-aware feature alignment for domain adaptive object detector,'' in
  \emph{European Conference on Computer Vision}.\hskip 1em plus 0.5em minus
  0.4em\relax Springer, 2020, pp. 733--748.

\bibitem{ATF}
Z.~He and L.~Zhang, ``Domain adaptive object detection via asymmetric tri-way
  faster-rcnn,'' in \emph{Computer Vision--ECCV 2020: 16th European Conference,
  Glasgow, UK, August 23--28, 2020, Proceedings, Part XXIV 16}.\hskip 1em plus
  0.5em minus 0.4em\relax Springer, 2020, pp. 309--324.

\bibitem{sindagi2019prior}
V.~A. Sindagi, P.~Oza, R.~Yasarla, and V.~M. Patel, ``Prior-based domain
  adaptive object detection for hazy and rainy conditions,'' in \emph{European
  Conference on Computer Vision}.\hskip 1em plus 0.5em minus 0.4em\relax
  Springer, 2020, pp. 763--780.

\bibitem{zhao2020collaborative}
G.~Zhao, G.~Li, R.~Xu, and L.~Lin, ``Collaborative training between region
  proposal localization and classification for domain adaptive object
  detection,'' in \emph{European Conference on Computer Vision}.\hskip 1em plus
  0.5em minus 0.4em\relax Springer, 2020, pp. 86--102.

\bibitem{ART-PCA}
Y.~Zheng, D.~Huang, S.~Liu, and Y.~Wang, ``Cross-domain object detection
  through coarse-to-fine feature adaptation,'' in \emph{Proceedings of the
  IEEE/CVF conference on computer vision and pattern recognition}, 2020, pp.
  13\,766--13\,775.

\bibitem{PIT}
Q.~Gu, Q.~Zhou, M.~Xu, Z.~Feng, G.~Cheng, X.~Lu, J.~Shi, and L.~Ma, ``Pit:
  Position-invariant transform for cross-fov domain adaptation,'' in
  \emph{Proceedings of the IEEE/CVF International Conference on Computer
  Vision}, 2021, pp. 8761--8770.

\bibitem{ASA}
W.~Zhou, Y.~Wang, J.~Chu, J.~Yang, X.~Bai, and Y.~Xu, ``Affinity space
  adaptation for semantic segmentation across domains,'' \emph{IEEE
  Transactions on Image Processing}, vol.~30, pp. 2549--2561, 2020.

\bibitem{CLANv2}
Y.~Luo, P.~Liu, L.~Zheng, T.~Guan, J.~Yu, and Y.~Yang, ``Category-level
  adversarial adaptation for semantic segmentation using purified features,''
  \emph{IEEE Transactions on Pattern Analysis and Machine Intelligence}, pp.
  1--1, 2021.

\bibitem{CyCADA}
J.~Hoffman, E.~Tzeng, T.~Park, J.-Y. Zhu, P.~Isola, K.~Saenko, A.~Efros, and
  T.~Darrell, ``Cycada: Cycle-consistent adversarial domain adaptation,'' in
  \emph{International conference on machine learning}.\hskip 1em plus 0.5em
  minus 0.4em\relax PMLR, 2018, pp. 1989--1998.

\bibitem{AdaptSegNet}
Y.-H. Tsai, W.-C. Hung, S.~Schulter, K.~Sohn, M.-H. Yang, and M.~Chandraker,
  ``Learning to adapt structured output space for semantic segmentation,'' in
  \emph{Proceedings of the IEEE conference on computer vision and pattern
  recognition}, 2018, pp. 7472--7481.

\bibitem{BDL}
Y.~Li, L.~Yuan, and N.~Vasconcelos, ``Bidirectional learning for domain
  adaptation of semantic segmentation,'' in \emph{Proceedings of the IEEE/CVF
  Conference on Computer Vision and Pattern Recognition}, 2019, pp. 6936--6945.

\bibitem{LTIR}
M.~Kim and H.~Byun, ``Learning texture invariant representation for domain
  adaptation of semantic segmentation,'' in \emph{Proceedings of the IEEE/CVF
  Conference on Computer Vision and Pattern Recognition}, 2020, pp.
  12\,975--12\,984.

\bibitem{CLAN}
Y.~Luo, L.~Zheng, T.~Guan, J.~Yu, and Y.~Yang, ``Taking a closer look at domain
  shift: Category-level adversaries for semantics consistent domain
  adaptation,'' in \emph{Proceedings of the IEEE/CVF Conference on Computer
  Vision and Pattern Recognition}, 2019, pp. 2507--2516.

\bibitem{SIM}
Z.~Wang, M.~Yu, Y.~Wei, R.~Feris, J.~Xiong, W.-m. Hwu, T.~S. Huang, and H.~Shi,
  ``Differential treatment for stuff and things: A simple unsupervised domain
  adaptation method for semantic segmentation,'' in \emph{Proceedings of the
  IEEE/CVF Conference on Computer Vision and Pattern Recognition}, 2020, pp.
  12\,635--12\,644.

\bibitem{DISE}
W.-L. Chang, H.-P. Wang, W.-H. Peng, and W.-C. Chiu, ``All about structure:
  Adapting structural information across domains for boosting semantic
  segmentation,'' in \emph{Proceedings of the IEEE/CVF Conference on Computer
  Vision and Pattern Recognition}, 2019, pp. 1900--1909.

\bibitem{FDA}
Y.~Yang and S.~Soatto, ``Fda: Fourier domain adaptation for semantic
  segmentation,'' in \emph{Proceedings of the IEEE/CVF Conference on Computer
  Vision and Pattern Recognition}, 2020, pp. 4085--4095.

\bibitem{STAR}
Z.~Lu, Y.~Yang, X.~Zhu, C.~Liu, Y.-Z. Song, and T.~Xiang, ``Stochastic
  classifiers for unsupervised domain adaptation,'' in \emph{Proceedings of the
  IEEE/CVF Conference on Computer Vision and Pattern Recognition}, 2020, pp.
  9111--9120.

\bibitem{PCEDA}
Y.~Yang, D.~Lao, G.~Sundaramoorthi, and S.~Soatto, ``Phase consistent
  ecological domain adaptation,'' in \emph{Proceedings of the IEEE/CVF
  Conference on Computer Vision and Pattern Recognition}, 2020, pp. 9011--9020.

\bibitem{IntraDA}
F.~Pan, I.~Shin, F.~Rameau, S.~Lee, and I.~S. Kweon, ``Unsupervised
  intra-domain adaptation for semantic segmentation through self-supervision,''
  in \emph{Unsupervised Intra-domain Adaptation for Semantic Segmentation
  through Self-Supervision}, 2020, pp. 3764--3773.

\bibitem{APODA}
J.~Yang, R.~Xu, R.~Li, X.~Qi, X.~Shen, G.~Li, and L.~Lin, ``An adversarial
  perturbation oriented domain adaptation approach for semantic segmentation,''
  in \emph{Proceedings of the AAAI Conference on Artificial Intelligence},
  vol.~34, no.~07, 2020, pp. 12\,613--12\,620.

\bibitem{FADA}
H.~Wang, T.~Shen, W.~Zhang, L.~Duan, and T.~Mei, ``Classes matter: {A}
  fine-grained adversarial approach to cross-domain semantic segmentation,'' in
  \emph{European conference on computer vision}, vol. 12359.\hskip 1em plus
  0.5em minus 0.4em\relax Springer, 2020, pp. 642--659.

\bibitem{DADA}
T.~Vu, H.~Jain, M.~Bucher, M.~Cord, and P.~P{\'{e}}rez, ``{DADA:} depth-aware
  domain adaptation in semantic segmentation,'' in \emph{Proceedings of the
  IEEE/CVF International Conference on Computer Vision}, 2019, pp. 7363--7372.

\bibitem{choi2019self}
J.~Choi, T.~Kim, and C.~Kim, ``Self-ensembling with gan-based data augmentation
  for domain adaptation in semantic segmentation,'' in \emph{Proceedings of the
  IEEE/CVF International Conference on Computer Vision}, 2019, pp. 6830--6840.

\bibitem{SIBAN}
Y.~Luo, P.~Liu, T.~Guan, J.~Yu, and Y.~Yang, ``Significance-aware information
  bottleneck for domain adaptive semantic segmentation,'' in \emph{Proceedings
  of the IEEE/CVF International Conference on Computer Vision}, 2019, pp.
  6778--6787.

\bibitem{CRST}
Y.~Zou, Z.~Yu, X.~Liu, B.~Kumar, and J.~Wang, ``Confidence regularized
  self-training,'' in \emph{Proceedings of the IEEE/CVF International
  Conference on Computer Vision}, 2019, pp. 5982--5991.

\bibitem{CBST}
Y.~Zou, Z.~Yu, B.~Kumar, and J.~Wang, ``Unsupervised domain adaptation for
  semantic segmentation via class-balanced self-training,'' in
  \emph{Proceedings of the European conference on computer vision (ECCV)},
  2018, pp. 289--305.

\bibitem{Conservative_loss}
X.~Zhu, H.~Zhou, C.~Yang, J.~Shi, and D.~Lin, ``Penalizing top performers:
  Conservative loss for semantic segmentation adaptation,'' in
  \emph{Proceedings of the European Conference on Computer Vision (ECCV)},
  2018, pp. 568--583.

\bibitem{AdaptPatch}
Y.-H. Tsai, K.~Sohn, S.~Schulter, and M.~Chandraker, ``Domain adaptation for
  structured output via discriminative patch representations,'' in
  \emph{Proceedings of the IEEE/CVF International Conference on Computer
  Vision}, 2019.

\bibitem{guo2021label}
S.~Guo, Q.~Zhou, Y.~Zhou, Q.~Gu, J.~Tang, Z.~Feng, and L.~Ma, ``Label-free
  regional consistency for image-to-image translation,'' in \emph{2021 IEEE
  International Conference on Multimedia and Expo (ICME)}.\hskip 1em plus 0.5em
  minus 0.4em\relax IEEE, 2021, pp. 1--6.

\bibitem{li2017domain}
W.~Li, Z.~Xu, D.~Xu, D.~Dai, and L.~Van~Gool, ``Domain generalization and
  adaptation using low rank exemplar svms,'' \emph{IEEE transactions on pattern
  analysis and machine intelligence}, vol.~40, no.~5, pp. 1114--1127, 2017.

\bibitem{ghifary2016scatter}
M.~Ghifary, D.~Balduzzi, W.~B. Kleijn, and M.~Zhang, ``Scatter component
  analysis: A unified framework for domain adaptation and domain
  generalization,'' \emph{IEEE transactions on pattern analysis and machine
  intelligence}, vol.~39, no.~7, pp. 1414--1430, 2016.

\bibitem{cordts2016cityscapes}
M.~Cordts, M.~Omran, S.~Ramos, T.~Rehfeld, M.~Enzweiler, R.~Benenson,
  U.~Franke, S.~Roth, and B.~Schiele, ``The cityscapes dataset for semantic
  urban scene understanding,'' in \emph{Proc. CVPR}, 2016, pp. 3213--3223.

\bibitem{FoggyCity}
C.~Sakaridis, D.~Dai, and L.~Van~Gool, ``Semantic foggy scene understanding
  with synthetic data,'' \emph{International Journal of Computer Vision}, vol.
  126, no.~9, pp. 973--992, 2018.

\bibitem{RTTS}
B.~Li, W.~Ren, D.~Fu, D.~Tao, D.~Feng, W.~Zeng, and Z.~Wang, ``Benchmarking
  single-image dehazing and beyond,'' \emph{IEEE Transactions on Image
  Processing}, vol.~28, no.~1, pp. 492--505, 2018.

\bibitem{FoggyDriving}
C.~Sakaridis, D.~Dai, S.~Hecker, and L.~Van~Gool, ``Model adaptation with
  synthetic and real data for semantic dense foggy scene understanding,'' in
  \emph{Proceedings of the European Conference on Computer Vision (ECCV)},
  2018, pp. 687--704.

\bibitem{rainy_city}
X.~Hu, C.-W. Fu, L.~Zhu, and P.-A. Heng, ``Depth-attentional features for
  single-image rain removal,'' in \emph{Proceedings of the IEEE/CVF Conference
  on Computer Vision and Pattern Recognition}, 2019, pp. 8022--8031.

\bibitem{VKITTI}
A.~Gaidon, Q.~Wang, Y.~Cabon, and E.~Vig, ``Virtual worlds as proxy for
  multi-object tracking analysis,'' in \emph{Proceedings of the IEEE conference
  on computer vision and pattern recognition}, 2016, pp. 4340--4349.

\bibitem{kITTI}
A.~Geiger, P.~Lenz, C.~Stiller, and R.~Urtasun, ``Vision meets robotics: The
  kitti dataset,'' \emph{IJR}, vol.~32, no.~11, pp. 1231--1237, 2013.

\bibitem{sim10k}
M.~Johnson-Roberson, C.~Barto, R.~Mehta, S.~N. Sridhar, K.~Rosaen, and
  R.~Vasudevan, ``Driving in the matrix: Can virtual worlds replace
  human-generated annotations for real world tasks?'' in \emph{2017 IEEE
  International Conference on Robotics and Automation (ICRA)}.\hskip 1em plus
  0.5em minus 0.4em\relax IEEE, 2017, pp. 746--753.

\bibitem{zhang2020unsupervised}
Y.~Zhang, B.~Deng, H.~Tang, L.~Zhang, and K.~Jia, ``Unsupervised multi-class
  domain adaptation: Theory, algorithms, and practice,'' \emph{IEEE
  Transactions on Pattern Analysis and Machine Intelligence}, 2020.

\bibitem{mancini2019inferring}
M.~Mancini, L.~Porzi, S.~R. Bulo, B.~Caputo, and E.~Ricci, ``Inferring latent
  domains for unsupervised deep domain adaptation,'' \emph{IEEE transactions on
  pattern analysis and machine intelligence}, 2019.

\bibitem{kouw2019review}
W.~M. Kouw and M.~Loog, ``A review of domain adaptation without target
  labels,'' \emph{IEEE transactions on pattern analysis and machine
  intelligence}, vol.~43, no.~3, pp. 766--785, 2019.

\bibitem{8943120}
W.~Zhang, D.~Xu, W.~Ouyang, and W.~Li, ``Self-paced collaborative and
  adversarial network for unsupervised domain adaptation,'' \emph{IEEE
  Transactions on Pattern Analysis and Machine Intelligence}, vol.~43, no.~6,
  pp. 2047--2061, 2021.

\bibitem{li2020deep}
S.~Li, C.~H. Liu, Q.~Lin, Q.~Wen, L.~Su, G.~Huang, and Z.~Ding, ``Deep residual
  correction network for partial domain adaptation,'' \emph{IEEE transactions
  on pattern analysis and machine intelligence}, 2020.

\bibitem{li2020maximum}
J.~Li, E.~Chen, Z.~Ding, L.~Zhu, K.~Lu, and H.~T. Shen, ``Maximum density
  divergence for domain adaptation,'' \emph{IEEE transactions on pattern
  analysis and machine intelligence}, 2020.

\bibitem{rozantsev2018beyond}
A.~Rozantsev, M.~Salzmann, and P.~Fua, ``Beyond sharing weights for deep domain
  adaptation,'' \emph{IEEE transactions on pattern analysis and machine
  intelligence}, vol.~41, no.~4, pp. 801--814, 2018.

\bibitem{liang2018aggregating}
J.~Liang, R.~He, Z.~Sun, and T.~Tan, ``Aggregating randomized
  clustering-promoting invariant projections for domain adaptation,''
  \emph{IEEE transactions on pattern analysis and machine intelligence},
  vol.~41, no.~5, pp. 1027--1042, 2018.

\bibitem{courty2016optimal}
N.~Courty, R.~Flamary, D.~Tuia, and A.~Rakotomamonjy, ``Optimal transport for
  domain adaptation,'' \emph{IEEE transactions on pattern analysis and machine
  intelligence}, vol.~39, no.~9, pp. 1853--1865, 2016.

\bibitem{DANN}
Y.~Ganin and V.~Lempitsky, ``Unsupervised domain adaptation by
  backpropagation,'' in \emph{International conference on machine learning},
  2015, pp. 1180--1189.

\bibitem{zhang2019curriculum}
Y.~Zhang, P.~David, H.~Foroosh, and B.~Gong, ``A curriculum domain adaptation
  approach to the semantic segmentation of urban scenes,'' \emph{IEEE
  transactions on pattern analysis and machine intelligence}, vol.~42, no.~8,
  pp. 1823--1841, 2019.

\bibitem{luo2021category}
Y.~Luo, P.~Liu, L.~Zheng, T.~Guan, J.~Yu, and Y.~Yang, ``Category-level
  adversarial adaptation for semantic segmentation using purified features,''
  \emph{IEEE Transactions on Pattern Analysis and Machine Intelligence}, 2021.

\bibitem{sakaridis2020map}
C.~Sakaridis, D.~Dai, and L.~Van~Gool, ``Map-guided curriculum domain
  adaptation and uncertainty-aware evaluation for semantic nighttime image
  segmentation,'' \emph{IEEE Transactions on Pattern Analysis and Machine
  Intelligence}, pp. 1--1, 2020.

\bibitem{DDMRL}
T.~Kim, M.~Jeong, S.~Kim, S.~Choi, and C.~Kim, ``Diversify and match: A domain
  adaptive representation learning paradigm for object detection,'' in
  \emph{Proceedings of the IEEE/CVF Conference on Computer Vision and Pattern
  Recognition}, 2019, pp. 12\,456--12\,465.

\bibitem{GAN}
I.~Goodfellow, J.~Pouget-Abadie, M.~Mirza, B.~Xu, D.~Warde-Farley, S.~Ozair,
  A.~Courville, and Y.~Bengio, ``Generative adversarial nets,'' vol.~27, 2014.

\bibitem{DRPC}
X.~Yue, Y.~Zhang, S.~Zhao, A.~Sangiovanni-Vincentelli, K.~Keutzer, and B.~Gong,
  ``Domain randomization and pyramid consistency: Simulation-to-real
  generalization without accessing target domain data,'' in \emph{Proceedings
  of the IEEE/CVF International Conference on Computer Vision}, 2019, pp.
  2100--2110.

\bibitem{huang2018multimodal}
X.~Huang, M.-Y. Liu, S.~Belongie, and J.~Kautz, ``Multimodal unsupervised
  image-to-image translation,'' in \emph{Proceedings of the European conference
  on computer vision}, 2018, pp. 172--189.

\bibitem{lee2018diverse}
H.-Y. Lee, H.-Y. Tseng, J.-B. Huang, M.~Singh, and M.-H. Yang, ``Diverse
  image-to-image translation via disentangled representations,'' in
  \emph{Proceedings of the European conference on computer vision (ECCV)},
  2018, pp. 35--51.

\bibitem{ridgeway2018learning}
K.~Ridgeway and M.~C. Mozer, ``Learning deep disentangled embeddings with the
  f-statistic loss,'' 2018.

\bibitem{scott2018adapted}
T.~R. Scott, K.~Ridgeway, and M.~C. Mozer, ``Adapted deep embeddings: A
  synthesis of methods for $ k $-shot inductive transfer learning,'' 2018.

\bibitem{liu2018detach}
Y.-C. Liu, Y.-Y. Yeh, T.-C. Fu, S.-D. Wang, W.-C. Chiu, and Y.-C.~F. Wang,
  ``Detach and adapt: Learning cross-domain disentangled deep representation,''
  in \emph{Proceedings of the IEEE Conference on Computer Vision and Pattern
  Recognition}, 2018, pp. 8867--8876.

\bibitem{IID}
A.~Wu, Y.~Han, L.~Zhu, and Y.~Yang, ``Instance-invariant domain adaptive object
  detection via progressive disentanglement,'' \emph{IEEE Transactions on
  Pattern Analysis and Machine Intelligence}, pp. 1--1, 2021.

\bibitem{mask-rcnn}
K.~He, G.~Gkioxari, P.~Doll{\'a}r, and R.~Girshick, ``Mask r-cnn,'' in
  \emph{Proceedings of the IEEE international conference on computer vision},
  2017, pp. 2961--2969.

\bibitem{CIDA}
H.~Wang, H.~He, and D.~Katabi, ``Continuously indexed domain adaptation,'' in
  \emph{The International Conference on Machine Learning}, 2020.

\bibitem{ren2015faster}
S.~Ren, K.~He, R.~Girshick, and J.~Sun, ``Faster r-cnn: Towards real-time
  object detection with region proposal networks,'' vol.~28, 2015, pp. 91--99.

\bibitem{zhang2021rpn}
Y.~Zhang, Z.~Wang, and Y.~Mao, ``Rpn prototype alignment for domain adaptive
  object detector,'' in \emph{Proceedings of the IEEE/CVF Conference on
  Computer Vision and Pattern Recognition}, 2021, pp. 12\,425--12\,434.

\bibitem{vs2021mega}
V.~VS, V.~Gupta, P.~Oza, V.~A. Sindagi, and V.~M. Patel, ``Mega-cda: Memory
  guided attention for category-aware unsupervised domain adaptive object
  detection,'' in \emph{Proceedings of the IEEE/CVF Conference on Computer
  Vision and Pattern Recognition}, 2021, pp. 4516--4526.

\bibitem{deng2021unbiased}
J.~Deng, W.~Li, Y.~Chen, and L.~Duan, ``Unbiased mean teacher for cross-domain
  object detection,'' in \emph{Proceedings of the IEEE/CVF Conference on
  Computer Vision and Pattern Recognition}, 2021, pp. 4091--4101.

\bibitem{wu2021vector}
A.~Wu, R.~Liu, Y.~Han, L.~Zhu, and Y.~Yang, ``Vector-decomposed disentanglement
  for domain-invariant object detection,'' in \emph{Proceedings of the IEEE/CVF
  International Conference on Computer Vision}, 2021, pp. 9342--9351.

\bibitem{MTOR}
Q.~Cai, Y.~Pan, C.-W. Ngo, X.~Tian, L.~Duan, and T.~Yao, ``Exploring object
  relation in mean teacher for cross-domain detection,'' in \emph{Proceedings
  of the IEEE/CVF Conference on Computer Vision and Pattern Recognition}, 2019,
  pp. 11\,457--11\,466.

\bibitem{zhang2018densely}
H.~Zhang and V.~M. Patel, ``Densely connected pyramid dehazing network,'' in
  \emph{Proceedings of the IEEE conference on computer vision and pattern
  recognition}, 2018, pp. 3194--3203.

\bibitem{liu2019griddehazenet}
X.~Liu, Y.~Ma, Z.~Shi, and J.~Chen, ``Griddehazenet: Attention-based
  multi-scale network for image dehazing,'' in \emph{Proceedings of the
  IEEE/CVF International Conference on Computer Vision}, 2019, pp. 7314--7323.

\bibitem{vgg}
K.~Simonyan and A.~Zisserman, ``Very deep convolutional networks for
  large-scale image recognition,'' \emph{arXiv preprint arXiv:1409.1556}, 2014.

\bibitem{he2016deep}
K.~He, X.~Zhang, S.~Ren, and J.~Sun, ``Deep residual learning for image
  recognition,'' in \emph{Proceedings of the IEEE conference on computer vision
  and pattern recognition}, 2016, pp. 770--778.

\bibitem{deng2009imagenet}
J.~Deng, W.~Dong, R.~Socher, L.-J. Li, K.~Li, and L.~Fei-Fei, ``Imagenet: A
  large-scale hierarchical image database,'' in \emph{2009 IEEE conference on
  computer vision and pattern recognition}.\hskip 1em plus 0.5em minus
  0.4em\relax Ieee, 2009, pp. 248--255.

\end{thebibliography}

\vspace{-1cm}

\end{document}